\documentclass[lettersize,journal]{IEEEtran}
\usepackage{amsmath,amsfonts}
\usepackage{algorithm}
\usepackage{array}
\usepackage[caption=true,font=normalsize,labelfont=sf,textfont=sf]{subfig} 
\usepackage{textcomp}
\usepackage{stfloats}
\usepackage{url}
\usepackage{verbatim}
\usepackage{graphicx}
\usepackage{cite}
\usepackage{xcolor}
\usepackage{booktabs}
\usepackage{graphicx}
\usepackage{algpseudocode}
\usepackage{lipsum}
\usepackage{etoolbox}
\usepackage{makecell}
\usepackage[
  colorlinks=true,
  linkcolor=red,
  citecolor=green,
  urlcolor=magenta
]{hyperref}
\begin{document}

\title{A Theory of Shape Reconstruction from \\ Heat Conduction and Shading}

\author{Akihiko~Oharazawa,
        Sriram~Narayanan,
        Mani~Ramanagopal,
        and~Srinivasa~G.~Narasimhan
}



\maketitle

\begin{abstract}
Shape from shading using a single image of a Lambertian surface is inherently ambiguous. When the light source direction is known, the surface normal estimation has a cone-ambiguity, which worsens when the source is unknown. Recently, shape from heat conduction has emerged as an approach that leverages heat transport equations to estimate the Shape Laplacian operator, an intrinsic measure of shape. However, deriving surface normals from the Laplacian operator encounters a local binary convex/concave ambiguity. Our contribution introduces a novel theory to resolve these local shape ambiguities (excluding a few degeneracies) without relying on priors like smoothness, by combining the cues from shading and heat conduction. Our method ensures the mathematical constraints of both shading and the Laplacian are satisfied simultaneously, even with an unknown light source. We validate our theory through simulations of complex shapes and analyze its performance in the presence of noise, as well as on a noisy single thermal video of real-world objects with complex shapes and material properties, including varying albedo. This paper extends the algorithm and results presented in our prior work \cite{Oharazawa_2025_ICCP}. Project page can be found at \url{https://shape-from-heat-and-shading.github.io/}.
\end{abstract}

\section{Introduction}
\IEEEPARstart{S}{hading} has long been a crucial cue for shape estimation in computer vision. Woodham \cite{Woodham_1979} introduced photometric stereo, a technique where shape and albedo are derived from images of a Lambertian surface illuminated by at least three known distant light sources. When the directions of these sources are unknown, Belhumeur et al. \cite{belhumeur1999bas} demonstrated that shape, albedo, and lighting can be estimated up to three degrees of freedom, known as the Generalized Bas-Relief (GBR) ambiguity. Subsequent works enhance the realism of the appearance model by modeling interreflections \cite{chandraker2005reflections} and inverse-square fall-off \cite{Prados_2005} to resolve the GBR ambiguity.

Even with a known source direction, when only a single shading image of a Lambertian surface is available, the estimation of the surface normal is highly ambiguous --- any normal along a cone around the source vector satisfies the shading/irradiance term ($n.s$). And if the source direction is unknown, the ambiguity is worse. Thus, historically, many approaches exploit additional shape (known boundary, smoothness, etc.) \cite{Horn_1970} and/or lighting priors \cite{barron2014shape} to overcome these ambiguities. Today, deep learning models recover depth maps from a single image based on supervised pre-training on large datasets of captured images and ground truth 3D models \cite{han2024multistable}. However, these models may struggle to generalize when the scene conditions differ significantly from the training data.

\begin{figure}
    \centering
    \includegraphics[width=1.02\linewidth, trim={1.2cm 15cm 9.4cm 1.2cm}]{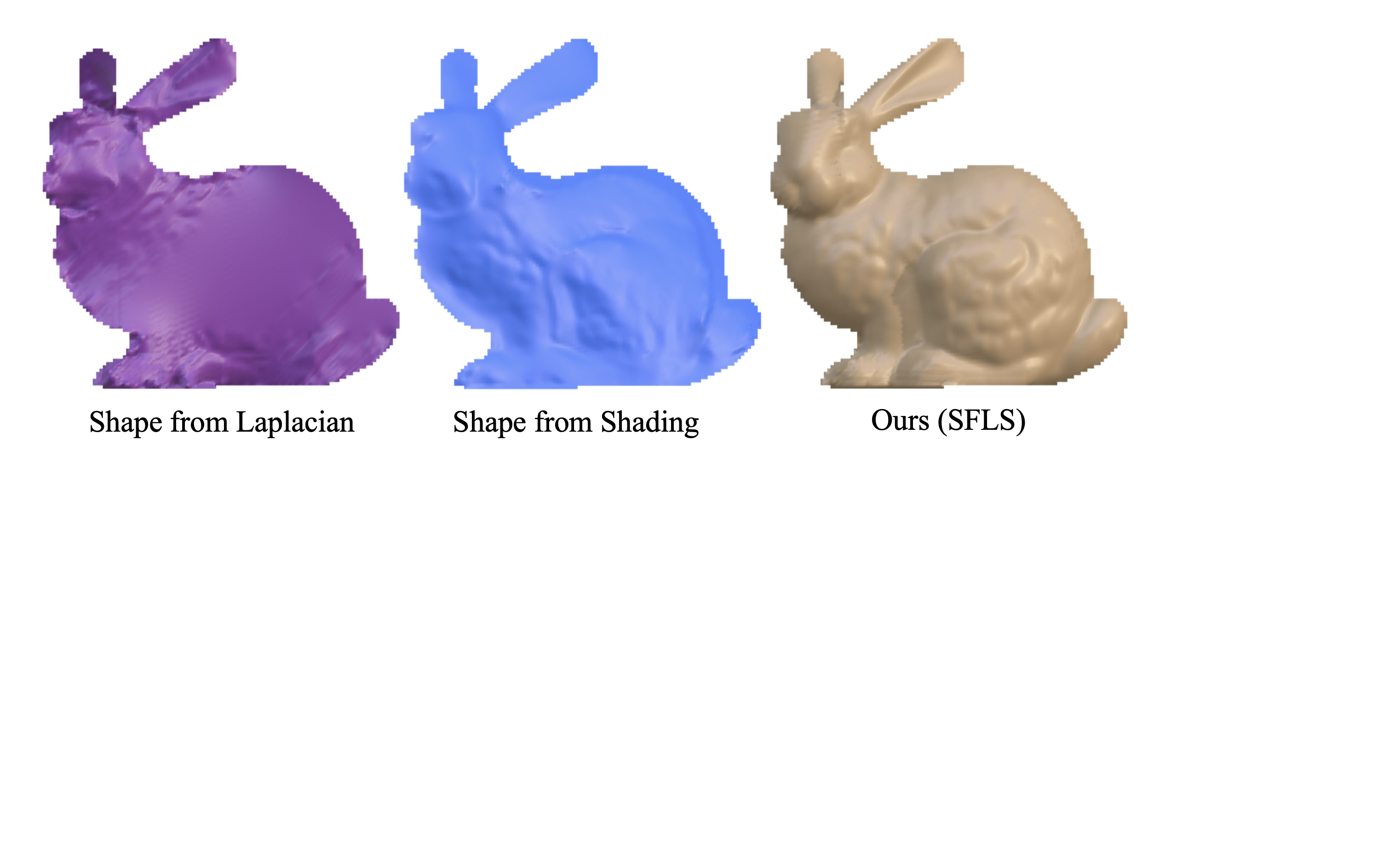}    
        \caption{Reconstructed shapes using estimation techniques based on Laplacian cues, shading cues, or both. While Laplacian cues constrain the local curvature around a pixel, it is invariant to convex/concave deformations. On the other hand, shading cues constrain the normal at each pixel to lie on a cone around the source direction, but do not constrain neighboring normals. This paper combines these complementary cues to resolve ambiguities while enabling faithful 3D reconstructions.}
    \label{fig:teaser}
    \vspace{-\intextsep}
\end{figure}

Recently, thermal heat conduction has been proposed as a strong cue of surface shape \cite{Narayanan_2024_ECCV}. From a single thermal video of an object undergoing either heating or cooling, the authors estimate the Shape Laplacian-Beltrami operator. This operator is an intrinsic representation of shape (as opposed to extrinsic surface normals or depths), and the approach works under arbitrary and unknown lighting and even for non-Lambertian surfaces. But computing normals or depth from the Laplacian operator suffers from a local binary convex/concave ambiguity. The authors suggest capturing 2-4 thermal videos under different lighting directions to resolve this ambiguity~\cite{Narayanan_2024_ECCV}.

Our key contribution lies in developing a novel theory to resolve the ambiguities in shape recovery by combining two cues: heat conduction and shading, under an unknown but distant light source. Mathematically, we resolve the ambiguities in normal estimation by simultaneously satisfying the Shape Laplacian and the Cosine (or $n.s$) term in shading. As before, shading is obtained by capturing a visible spectrum image of the scene. Heat conduction is obtained using the thermal transient video of the scene that is illuminated (and hence heated) by the light source. For the purposes of our theory, we assume the surface albedos are either uniform or are estimated {\it a priori} using the visible-thermal intrinsic decomposition approach \cite{Ramanagopal_2024_CVPR}, \cite{yuan2025vt-intrinsic}.

Following \cite{Narayanan_2024_ECCV}, we model the Shape Laplacian using cotangent weights on a mesh surface. Then, we develop a mathematical relationship between these weights and the local vertex surface normals. We consider the case of known source direction first. Our method estimates normals by satisfying both constraints  --- the equations for cotangent weights and the cosine term for irradiance --- simultaneously, resolving ambiguities except in a few degenerate cases. Then, we consider the general case of an unknown source direction and develop an algorithm that uniquely estimates both the normals (see Fig. \ref{fig:teaser}) and source direction. 

Previous approaches use priors such as smoothness (including penalizing Laplacian or second derivatives of estimated normals, for instance) that often clash with the ``data'' terms and require setting a regularization hyperparameter that changes with iterations. But our shape Laplacian is actually a ``data term'' that provides the correct smoothness too without the need for any regularization parameters. 

In our prior work\cite{Oharazawa_2025_ICCP}, we proposed a sequential optimization algorithm applied in a pixel-wise manner. In this work, we formulate a simpler and faster (from hours to minutes) approach by recognizing that the analytical solution under locally planar approximation provides a good initialization even without any information about the light source and even in non-planar regions. Leveraging this, we formulate a single global optimization problem (after estimating source from confidently initialized surface normals), that estimates shape at all pixels simultaneously. To stabilize the optimization, we enforce the integrability constraint using a penalty term. 

We evaluate the proposed framework through both simulations and experiments on real-world data. In simulation, we evaluate the method on complex shapes and compare with methods relying solely on Laplacian cues, solely on shading cues, or a full analysis-by-synthesis approach. We further evaluate robustness under various non-idealities in simulation including different levels of noise in the input Laplacian and shading,  challenging appearance such as interreflections, non-Lambertian reflectance, and spatially varying albedo. Finally, experiments on multiple real objects captured with a thermal camera and a single light source show that the proposed method resolves the concave/convex ambiguity while accurately reconstructing surface geometry.
\section{Related Work}
Single-view shape recovery from shading has been studied for several decades and remains an active research topic in computer vision\cite{ikeuchi1981numerical, frankot1988method,lee1997shape, zhang1999shape}. Under the assumption of Lambertian reflectance, shape-from-shading is fundamentally ill-posed, especially when the light source is unknown. This is because the observed shading can be explained by a multidimensional space of possible surface normal \cite{heal2020lighting}. Over the years, a variety of techniques have been proposed to address this ill-posedness, including partial differential equation (PDE)\cite{Horn_1970}, optimization\cite{szeliski1990}, and local analysis or linearization approaches\cite{tsai1994}. However, these methods often suffer from practical issues such as convergence to local minima. As a result, many shape-from-shading methods either assume known illumination or introduce strong priors to constrain the solution space. For example, SIRFS \cite{barron2014shape} employs distinct lighting priors for natural versus laboratory environments and imposes surface normal priors around the occluding boundaries. Despite such advancements, fully resolving the ambiguity of shape-from-shading under the single-view and unknown-lighting scenario remains an open challenge.

Thermal imaging offers a promising avenue to alleviate some of these ambiguities. Because thermal cameras capture wavelengths outside the visible spectrum, they enable new applications that leverage material properties, temperature distributions, and radiative characteristics that are otherwise difficult to observe. Recent advances include pose estimation for humans or objects \cite{ThermalNLOS}, navigation \cite{sheinin2024projecting}, material classification \cite{MatClasROBIO} and the separation of reflected and emitted thermal radiation \cite{narayanan2025dual}. Moreover, using the heat conduction equation, researchers have estimated diffusivity and emissivity for planar objects \cite{TSF2023}. In \cite{Ramanagopal_2024_CVPR} and \cite{yuan2025vt-intrinsic}, the authors introduce the physical relationship between a visible image and a thermal video of the scene and demonstrate that intrinsic image decomposition becomes a well-posed problem. 

Several efforts have explored shape estimation via thermal imaging. Nagase et al. \cite{SFTR} proposed a method for 3D reconstruction using multispectral thermal radiation while accounting for air attenuation. In \cite{kitazawa2025shape} and \cite{Kushida_2025_ICCV}, the authors explored shape-from-polarization in the Long-Wave Infrared spectrum using polarized thermal radiation. Tanaka et al. \cite{TPS} introduced a photometric stereo approach using steady-state (equilibrium) thermal images. Because the heat conduction equation strongly depends on surface curvature and boundary conditions, the transient response of thermal diffusion can, in principle, reveal shape information. Narayanan et al. \cite{Narayanan_2024_ECCV} exploited the Laplacian operator from the heat equation to reconstruct surface shape, thereby highlighting the potential of time-resolved thermal observations. However, their approach still requires two or more light sources to resolve local concavity or convexity ambiguities. In contrast, our method assumes a single, uncalibrated light source and highlights the relationship between shading and Laplacian cues derived from heat conduction. We provide a framework for the simultaneous estimation of unknown light source direction and shape, and theoretically demonstrate that the mathematical constraints of both shading and Laplacian cues resolve local ambiguities.


\begin{figure}[t]
    \centering
    \includegraphics[width=1.00\linewidth, trim={3cm 7.2cm 1.5cm 5.8cm}, clip]{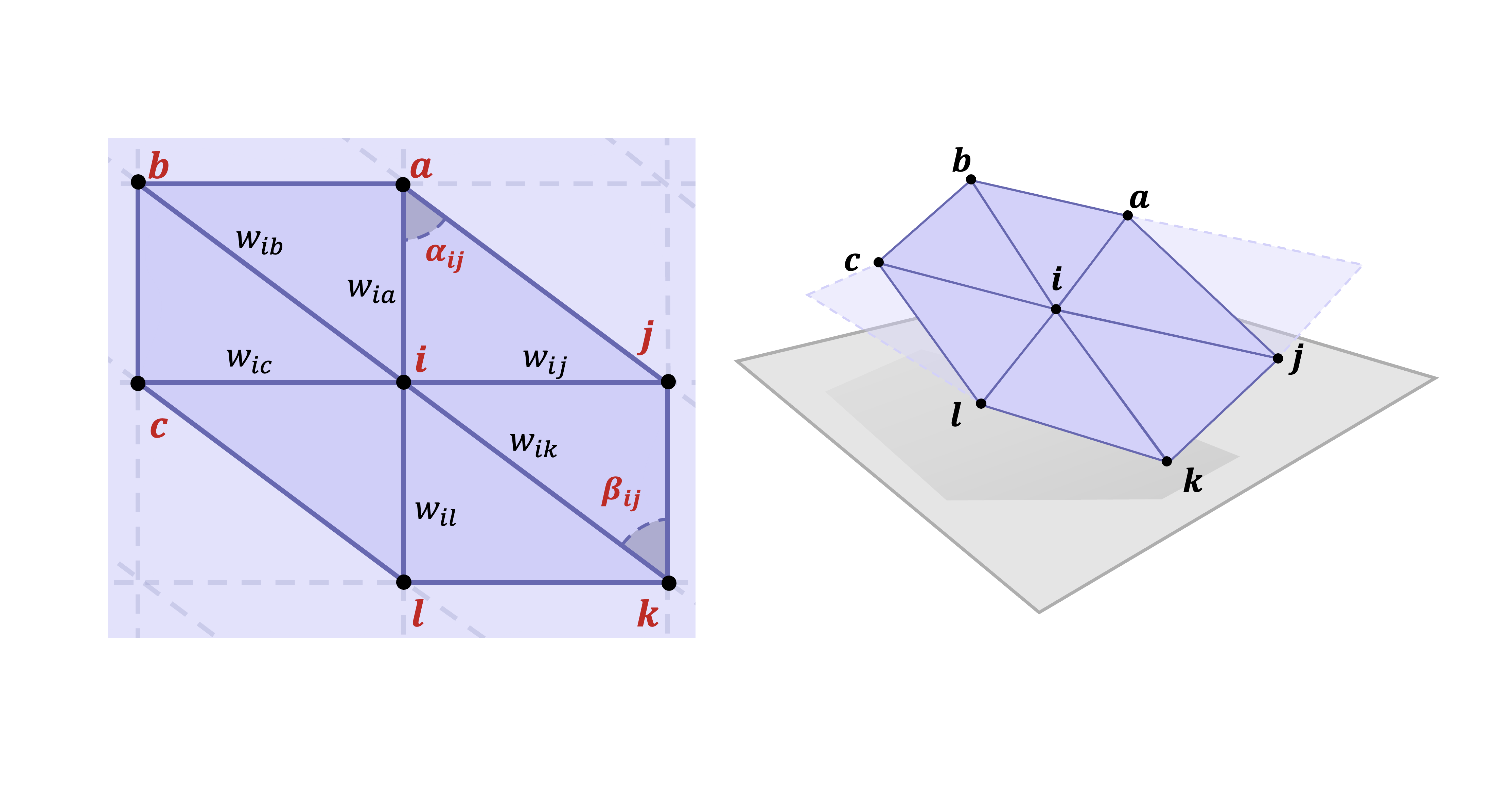}
    \caption{
    \textbf{Left:} A regular grid triangular mesh whose cotangent weights for any vertex $i$ depend on its neighboring six vertices.
    \textbf{Right:} Illustrate the local surface patch around vertex $i$, as an embedding of the triangular mesh in 3D space.}
    \label{fig:mesh_grid}
    \vspace{-\intextsep}
\end{figure}
\section{Problem Statement}
The objective of this work is to estimate the scene geometry (shape) and the light source direction ($\hat{\mathbf{s}}$) given only a single shading image ($\eta$) and the shape Laplacian ($\textrm{L}$). 
Similar to the original Shape-from-Shading (SfS) problem definition, we consider an orthographic camera and represent the shape using surface normals $(p_i,q_i)$ at each pixel $i$.
Unlike SfS, we do not know the light source direction and do not use an arbitrary smoothness term. 

Let $\Omega$ denote the unknown surface in 3D.
We discretize it into a mesh with $V$ vertices, with each vertex lying along the camera ray from a corresponding pixel.
We use a regular grid connectivity such that each vertex is connected to its six neighbors, as depicted in Fig.~\ref{fig:mesh_grid}.

The shading $\eta$ at a pixel $i$ can be written as
\begin{equation}
    \eta(i) = \frac{\vec{\mathbf{n}}(i)}{\|\vec{\mathbf{n}}(i)\|} \cdot \frac{\vec{\mathbf{s}}}{\|\vec{\mathbf{s}}\|} \;\; \textrm{where} \;\; \vec{\mathbf{n}}(i) = \begin{bmatrix}
        p_i \\ q_i \\ -1
    \end{bmatrix}, \, \vec{\mathbf{s}} = \begin{bmatrix}
        p_s \\ q_s \\ -1
    \end{bmatrix}.
    \label{eq:shading}
\end{equation}
For scenes with spatially varying albedo, one could use ~\cite{Ramanagopal_2024_CVPR} to recover shading from the visible image and thermal video. 
Specifically, each non-zero entry represents a weight \(w_{ij}\) between vertices \(i\) and \(j\) described as~\cite{Narayanan_2024_ECCV}: 
\begin{equation}
    \textrm{L}_{ij} = \begin{cases}
        w_{ij} = \dfrac{\cot \alpha_{ij} + \cot \beta_{ij}}{2} & \text{if } i \neq j\\
        -\sum_{j \in \mathcal{N}(i)} w_{ij} & \text{if } i = j
    \end{cases}
    \label{eq:cotan-lap}
\end{equation}
where \(\alpha_{ij}\) and \(\beta_{ij}\) are the angles of the two triangles adjacent to the edge \((i,j)\). 
These edge weights can be computed directly from vertex normals, as we will show in Sec.~\ref{subsec:L_to_pq}.

\section{Shape from Laplacian and Shading}
We explore how the intrinsic geometry captured by Laplacian can be combined with photometric cues from shading to recover surface normals more accurately and unambiguously. While Laplacian-based methods offer geometric structure, they suffer from local ambiguities in normal direction. Conversely, shape-from-shading provides photometric constraints but has its own set of ambiguities tied to lighting. In this section, we introduce a novel formulation that connects Laplacian weights directly to surface normals, allowing shading information to be integrated seamlessly. We also analyze the nature of ambiguities in each modality and show how their intersection leads to a unique and consistent surface reconstruction.

\subsection{Shape Laplacians using Vertex Normals}
\label{subsec:L_to_pq}

We introduce a novel parameterization of Laplacian weights in terms of vertex normals. The surface normal at a vertex can be expressed using depth gradients as $p = \frac{\partial z}{\partial x}$ and $q = \frac{\partial z}{\partial y}$. Discretizing these derivatives and using the notation illustrated in Fig.~\ref{fig:mesh_grid}, we can write:
\begin{equation} 
    p_i = \frac{z_j - z_i}{\delta_p}, \quad q_i = \frac{z_l - z_i}{\delta_p} 
\end{equation}
where $\delta_p$ is the pixel pitch and $z_i$ is the depth of vertex $i$.

The edge vectors connecting vertex $a$ to its neighbors can then be written in terms of normals:
\begin{equation} 
    \vec{ai} = (0, \delta_p, z_i - z_a) \quad \vec{aj} = (\delta_p, \delta_p, z_j - z_a) 
\end{equation}
Substituting depth differences in terms of $p$ and $q$:
\begin{equation}
    \begin{aligned}
        z_i - z_a &= q_a \cdot \delta_p \\
        z_j - z_a &= (p_i + q_a) \cdot \delta_p \\
    \end{aligned}
\end{equation}
Using these, we compute the cotangent of the angle $\alpha_{ij}$ between edges $\vec{ai}$ and $\vec{aj}$:
\begin{equation} 
    \cot \alpha_{ij} = \frac{|\vec{ai} \cdot \vec{aj}|}{|\vec{ai} \times \vec{aj}|} 
    = \frac{|1 + p_i q_a + q_a^2|}{\sqrt{p_i^2 + q_a^2 +1}}
    \label{eq:cotan_normals} 
\end{equation}
Similarly,
\begin{equation} 
    \cot \beta_{ij} = \frac{|\vec{ki} \cdot \vec{kj}|}{|\vec{ki} \times \vec{kj}|} 
    = \frac{|1 + p_i q_j + q_j^2|}{\sqrt{p_i^2 + q_j^2 + 1}}
    \label{eq:cotan_normals_beta} 
\end{equation}
Therefore, one can write the weight $w_{ij}$ in $\textrm{L}$ from Eq. \ref{eq:cotan-lap} as,
\begin{equation}
    w_{ij} = \frac{1}{2}\Bigg(\frac{|1 + p_i q_a + q_a^2|}{\sqrt{p_i^2 + q_a^2 +1}} + \frac{|1 + p_i q_j + q_j^2|}{\sqrt{p_i^2 + q_j^2 + 1}}\Bigg).
    \label{eq:cotan-weight}
\end{equation}

This establishes a direct relationship between the cotangent weights and the surface vertex normals. The full set of equations for all six edge weights is provided in the Appendix.

\begin{figure}[t]
    \centering
    \includegraphics[width=\linewidth, trim={0cm 0cm 0cm 0cm}, clip]{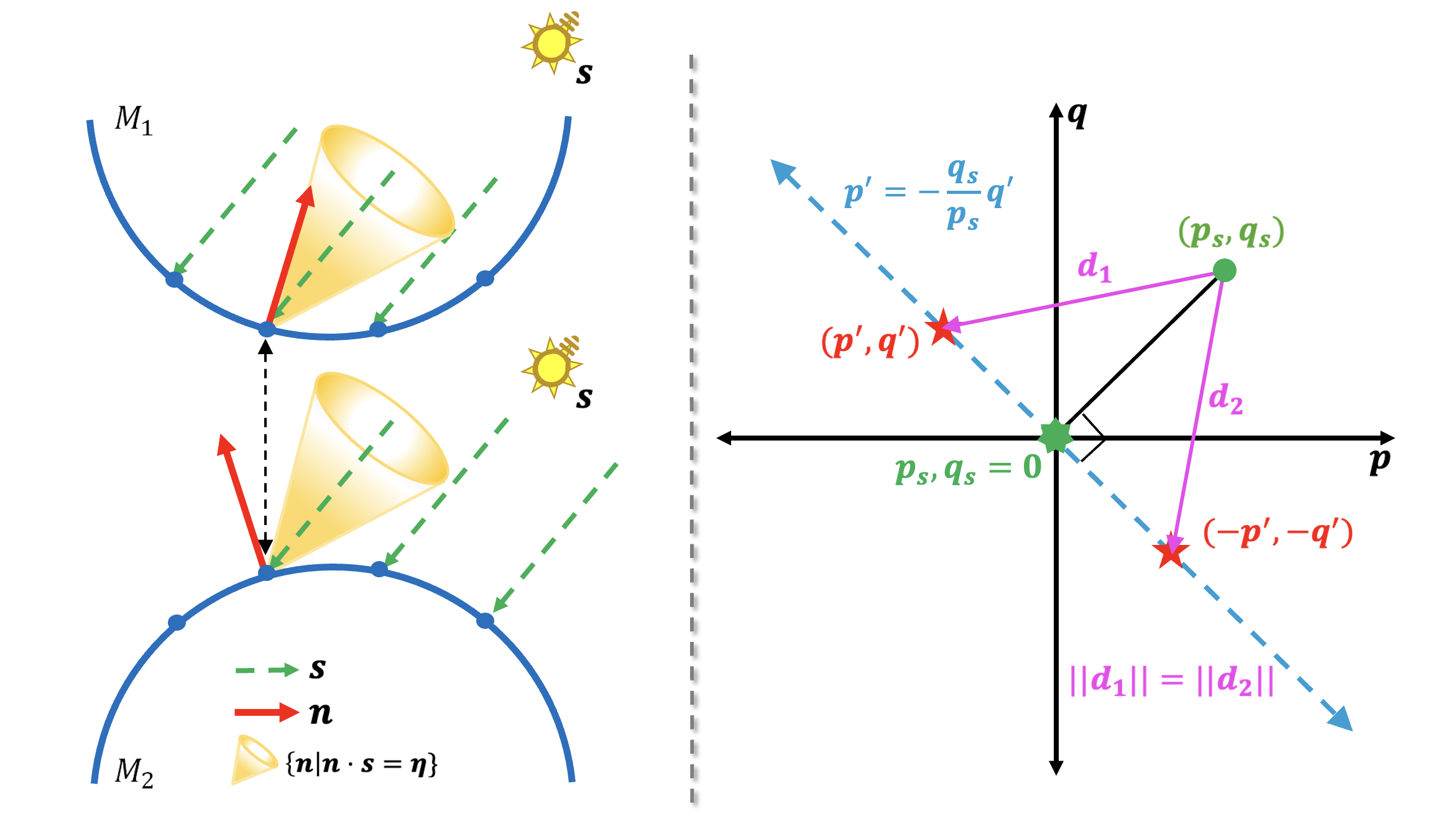}
    \caption{
    \textbf{Left:} Illustrates ambiguity in shape-from-shading, where a cone of normals yields identical shading, and a local binary ambiguity in shape-from-Laplacian, where different shapes with the same cotangent weights appear as mirror reflections in normal space. The true object shape is uniquely determined at the intersection of the shading cone and Laplacian normals. 
    \textbf{Right:} Demonstrates degenerate cases where the proposed shape-from-Laplacian-and-shading method fails.}
    \label{fig:ambiguity_exp}
    \vspace{-\intextsep}
\end{figure}

\subsection{Resolving Shape Ambiguities}
\label{ssec:resolveAmb}

In this section, we show that combining Laplacian-based and shading-based cues, under known distant illumination, resolves the shape ambiguity that arises when each is used independently. Specifically, we demonstrate that the intersection of their solution spaces leads to a unique normal direction at each point, except in a few degenerate cases. We begin by revisiting the types of ambiguities that arise when estimating shape from either Laplacian or shading cues alone, assuming the light source direction \((p_s, q_s)\) and shading values \(\eta\) are known. While we make this assumption here for clarity, the following section describes how the source direction can also be robustly estimated through our optimization framework.

\subsubsection{Ambiguities in Shape from Laplacian}

As shown in \cite{Narayanan_2024_ECCV}, the Laplace operator is intrinsic to a surface and invariant to rigid transformations. This means there are infinitely many 3D embeddings that share the same Laplacian. However, when vertices are constrained to lie along the camera rays, the number of possible embeddings reduces to 2 i.e. there is a local binary ambiguity in shape. 
This can be seen by flipping the sign of all vertex normals in a patch, i.e. substituting $(-p, -q)$ in place of $(p,q)$, in Eq.~\ref{eq:cotan-weight} which does not change $\textrm{L}$. 
This local binary ambiguity arises from isometric deformations and reflects a convex/concave uncertainty at each point.

\subsubsection{Ambiguities in Shape from Shading}

Shape-from-shading suffers from a different kind of ambiguity. Even when the albedo is constant and the light source direction $(p_s, q_s)$ is known, the surface normal is not uniquely determined. Instead, all normals that make the same angle with the light source produce the same shading. 
Mathematically, from Eq.~\ref{eq:shading}, this is written as, 
\begin{equation}
    \frac{pp_s + qq_s + 1}{\sqrt{p^2 + q^2 + 1} \sqrt{p_s^2 + q_s^2 + 1}} = \eta
    \label{eq:pq_eta}
\end{equation}
This defines a cone of possible normals around the light source direction, creating a continuous ambiguity distinct from the binary ambiguity of Laplacian-based methods.

\subsubsection{Intersection of Shading and Laplacian Ambiguities}
We seek a normal direction consistent with both the Laplacian constraint and the shading equation (Fig.~\ref{fig:ambiguity_exp}, left). If both Laplacian normals \((p,q)\) and \((-p,-q)\) satisfy Eq.~\ref{eq:pq_eta}, then:
\begin{align}
    pp_s + qq_s + 1 &= -pp_s - qq_s + 1 \\
    \Rightarrow \quad pp_s + qq_s &= 0
\end{align}
This yields the condition $pp_s = -qq_s$ corresponding to a narrow set of surface orientations in the normal space. This ambiguity persists only under special cases: (i) Fronto-parallel lighting: when $p_s = q_s = 0$, where both normal directions yield identical shading. (ii) Specific surface orientation: when $p = -\frac{q_s}{p_s} q$, representing a plane orthogonal to the bisector of $(p_s, q_s)$ and $(-p_s, -q_s)$.

Geometrically, these cases occur when the light direction $(p_s, q_s)$ lies equidistant between $(p, q)$ and its mirror reflection $(-p, -q)$ in the $pq$ space--i.e., along the line where shading distance to both normals is equal (see Fig. \ref{fig:ambiguity_exp}--right). In such cases, both candidate normals fall within the same shading cone defined by Eq.~\ref{eq:pq_eta}, making the ambiguity unresolvable.

However, these conditions occur only for some rare and isolated directions and typically do not persist across an entire surface. Thus, in general lighting conditions, combining Laplacian and shading cues yields a unique surface normal almost everywhere.

\section{Jointly Estimating Lighting and Shape}
\label{sec:joint_est}


In the previous section, we showed how ambiguities arising independently from shading and Laplacian constraints can be resolved when the light source direction is known. Building on this theoretical foundation, we now address the more general setting in which both the surface shape and a distant light source are unknown. In this section, we present a unified optimization framework that jointly estimates the uncalibrated light source direction and dense surface normals directly from shading and discrete Laplacian geometry.

Our approach follows a three-stage procedure as outlined in Algorithm \ref{alg:sfls} that progressively couples photometric and geometric constraints to arrive at a globally consistent solution. We begin by analytically approximating surface normals from local Laplacian structure under a near-planarity assumption. Rather than explicitly segmenting the surface into planar regions, we exploit characteristic symmetries in the cotangent Laplacian weights that arise when local surface variation is small. This yields reliable closed-form estimates of normal magnitudes (up to a sign ambiguity) across the surface, which serve as a robust initialization without requiring any prior knowledge of the lighting direction (Section~\ref{ssec:initPlanarNorm}).

Using these analytically initialized normals, we next estimate the unknown light source direction via a reverse photometric stereo formulation derived from the shading equation. Since both the Laplacian and shading constraints admit a global sign ambiguity, we resolve the relative sign consistency between surface normals and lighting through an alternating, RANSAC-based optimization that jointly refines the light source direction and the initialized normals until convergence (Section~\ref{ssec:SGN}).

Finally, given the estimated light source direction, we refine surface normals over the entire image domain to obtain a globally consistent shape estimate. This final stage enforces consistency with both the discrete cotangent Laplacian and the integrability of the normal field, ensuring that the recovered normals respect the underlying surface geometry while remaining consistent with observed shading (Section~\ref{ssec:SGSDir}). The resulting optimization yields dense surface normals that simultaneously satisfy photometric and geometric constraints, forming the basis for subsequent shape reconstruction.

The remainder of this section details each stage of the proposed framework, including analytical normal initialization from Laplacian structure, joint estimation of lighting and normals, and global refinement under Laplacian and integrability constraints.


\subsection{Analytical Normal Approximation under Local Planarity} 
\label{ssec:initPlanarNorm}

In our pixel-based mesh parameterization with connectivity shown in Fig. \ref{fig:mesh_grid} we can see that each vertex is connected to at most six neighbors. Under the assumption of local near-planarity, the Laplacian weights of diagonally opposite edges become approximately equal as
\begin{equation}
    w_{ij} \approx w_{ic}, \quad w_{ik} \approx w_{ib}, \quad w_{ia} \approx w_{il}.
    \label{eq:diagonal}
\end{equation}
This symmetry allows us to derive a closed-form approximation of the surface normals from the local Laplacian structure.
The Laplacian weights at each pixel $i$ can be expressed in terms of the unknown local surface normal $(p,q)$ as:
\begin{align}
    w_1 = \frac{w_{ia} + w_{il}}{2} &\approx \frac{p(p + q) + 1}{\sqrt{p^2 + q^2 + 1}},  \notag \\
    w_2 = \frac{w_{ij} + w_{ic}}{2} &\approx \frac{q(p + q) + 1}{\sqrt{p^2 + q^2 + 1}}, \label{eq:planar_weights} \\
    w_3 = \frac{w_{ik} + w_{ib}}{2} &\approx \frac{-pq}{\sqrt{p^2 + q^2 + 1}}. \notag    
\end{align}
Using the symmetry in the above equations, note that
\begin{equation}
    w_1 + w_2 + 2w_3 = \frac{p^2 + q^2 + 2}{\sqrt{p^2 + q^2 + 1}}.
    \label{eq:sum_w123}
\end{equation}
Taking $x = \sqrt{p^2 + q^2 + 1}$, where $x \geq 1$ by definition, we get
\begin{equation}
    \frac{x^2 + 1}{x} = w_1 + w_2 + 2 w_3.
    \label{eq:quad_eqn}
\end{equation}
We solve this quadratic equation in $x$ and pick the solution that is greater than or equal to 1. In Appendix, we show that the other solution is less than 1.   
To recover the values of $p$ and $q$, observe from \eqref{eq:planar_weights} that
\begin{equation}
    w_1 + w_3 = \frac{p^2 + 1}{x}.
\end{equation}
From this equation, we can compute $|p| = \sqrt{(w_1 + w_3) x - 1}$ and similarly, $|q| = \sqrt{(w_2 + w_3)x - 1}$.
Note that the sign of $w_3$ restricts the relative sign of $p,q$. Finally, the two solutions corresponding to the Shape Laplacians are,
\begin{equation}
    p=\sigma|p|,\quad
    q=-\sigma\, \mathrm{sgn}(w_3)\, |q|,\quad
    \sigma\in\{+1, -1\}
    \label{eq:mod_pq}
\end{equation}
where $\mathrm{sgn}(\cdot)$ denotes the sign function.
When $w_3 = 0$, the solutions simplify to either $p=0$ and $q=\pm\sqrt{w_2^2-1}$ or $p=\pm\sqrt{w_1^2-1}$ and $q=0$.

This formulation allows us to recover surface normals $\mathbf{n}_{\mathcal{V}} = \{ (p_i, q_i, -1)\  \forall i \in \mathcal{V}\}$ (up to $\sigma$) directly from Laplacian weights where $\mathcal{V}$ is the all vertices on the mesh.


            



\begin{algorithm}[t]
    \caption{Shape from Laplacian and Shading (SFLS)}
    \begin{algorithmic}[1]
        \Procedure{SFLS}{$\mathbf{L}, \boldsymbol{\eta}$}
            \State \texttt{// Init normal at every pixel}
            \State $\mathbf{n}^{0}_{\mathcal{V}} \gets \texttt{approxNormalsFromLaplacian}(\mathbf{L})$
            \State $\mathbf{n}_{\mathcal{V}} \gets \mathbf{n}^{0}_{\mathcal{V}}$

            \State \texttt{// Optimize }$\mathbf{\hat{s}}$\texttt{ and refine }$\mathbf{n}_{\mathcal{V}}$
            \While{$\|\Delta \hat{\mathbf{s}}\| > \epsilon_s$ \textbf{or} $\|\Delta \mathbf{n}_{\mathcal{V}}\| > \epsilon_n$}
                \State $\hat{\mathbf{s}} \gets \texttt{estimateSourceDir}(\mathcal{V}, \mathbf{n}_{\mathcal{V}}, \boldsymbol{\eta})$
                \State $\mathbf{n}_{\mathcal{V}} \gets \texttt{shapeGivenSource}(\mathcal{V}, \mathbf{n}_{\mathcal{V}}, \mathbf{L}, \hat{\mathbf{s}})$
            \EndWhile
            
            \State \Return $\mathbf{n}_{\mathcal{V}}$
        \EndProcedure
    \end{algorithmic}
    \label{alg:sfls}
\end{algorithm}

\subsection{Estimating Source Given Normals}
\label{ssec:SGN}

Once surface normals are known, we can estimate the lighting direction using a reverse photometric stereo approach, derived from the shading equation in Eq.~\ref{eq:pq_eta}. In matrix form, this estimation can be expressed as:
\begin{equation}
    \begin{bmatrix}
        \tilde{p}_0 & \tilde{q}_0 & \alpha_0\\
        \vdots & \vdots & \vdots \\
        \tilde{p}_N & \tilde{q}_N & \alpha_N
    \end{bmatrix} 
    \begin{bmatrix}
        \hat{p}_s \\ \hat{q}_s \\ \beta_s
    \end{bmatrix} = 
    \begin{bmatrix}
        \eta_0\\
        \vdots \\
        \eta_N
    \end{bmatrix}
    \label{eq:reverse_ps}
\end{equation}
Here, $(\hat{p}_s, \hat{q}_s, \beta_s)$ represents the estimated lighting direction, inferred from $N$ observed pixels. The term $\beta_s$ is a normalization factor. 

The entries $\tilde{p}$ and $\tilde{q}$ in the above matrix are normalized components of the surface normals, computed as $\tilde{p} = \alpha p$ and $\tilde{q} = \alpha q$, where $\alpha = 1/\sqrt{p^2 + q^2 + 1}$ ensures unit norm.

However, an inherent ambiguity arises because a surface normal $(\tilde{p}, \tilde{q})$ and its flipped version $(-\tilde{p}, -\tilde{q})$ yield identical shading under correspondingly flipped lighting direction $(-p_s, -q_s)$. 
At the same time, both versions of the surface normals satisfy the Laplacian weights. 
Here, the value of $(\tilde{p}, \tilde{q})$ corresponds to one of the two signed solutions from the previous subsection. 

To disambiguate the correct sign for the surface normals, we apply a RANSAC~\cite{Fischler_1981} based approach. 
We randomly pick a light source direction and then determine the sign for each surface normal which produces the lower shading error. 
Using these signs, we compute the optimal light source direction. 
We repeat until both the light source direction and signs of the normals are converged. The signs of the final light source direction $[p_s, q_s, -1]^T$ and surface normals $[p_i, q_i, -1]^T$ corresponds to one of two global solutions, with the other solution being $[-p_s, -q_s, -1]^T$ and $[-p_i, -q_i, -1]^T \forall i$. In both cases, the light source and camera are on the same side of the object. Our method enumerates both these global solutions. Without loss of generality, we pick one of the light source directions and proceed to estimate its corresponding shape.

\subsection{Estimating Shape Given Source Direction}
\label{ssec:SGSDir}

In this section, we estimate the surface normals across the entire object given an estimated source direction $\mathbf{\hat{s}}$ by enforcing both shading and Laplacian constraints. Section \ref{ssec:initPlanarNorm} provides approximate magnitudes of the surface normals, $|p|$ and $|q|$, across the entire object. Since Eq.~\ref{eq:mod_pq} yields two possible solutions, a sign ambiguity arises. 
To resolve this ambiguity, we evaluate both sign candidates in Eq. \ref{eq:pq_eta} and select the one that minimize the shading error with respect to the observed shading value $\eta$.
The candidate that minimizes the shading error is selected and assigned sign. 
For numerical stability, candidates where $p\approx0$, $q\approx0$, or $p = -\frac{q_s}{p_s} q$ are detected using a small threshold and excluded. 

Although the above formulation yields an initial estimate consistent with the shading constraint, it does not explicitly enforce global geometric consistency.
We then further refine the surface normals by enforcing both the Laplacian and integrability constraints. Specifically, we define an objective function over the set of all vertices $\mathcal{V}$ on the mesh. Let $\mathcal{N}(i)$ denote the set of neighboring vertices of vertex $i$, and $w_{ik}$ the known cotangent weights derived from the estimated surface normals. The Laplacian loss is defined as
\begin{equation}
    \mathcal{L}_{lap} = 
    \sum_{i \in \mathcal{V}}
    \sum_{k \in \mathcal{N}(i)}
    \left( w_{ik} - \hat{w}_{ik}(\mathbf{n}) \right)^2
\end{equation}
Here, $\hat{w}_{ik}(\mathbf{n})$ are the weights derived from the initial normal estimates. 
The integrability loss $\mathcal{L}_{int}$ enforces the consistency of the gradient field. For each $i \in \mathcal{V}$, we define
\begin{equation}
    \mathcal{L}_{int} =
    \sum_{i \in \mathcal{V}}
    \left((\partial_y p)_i - (\partial_x q)_i \right)^2
\end{equation}
The overall optimization problem is formulated as
\begin{equation}
    \mathbf{n}^* = \underset{\mathbf{n}}{\arg\min} \
    \lambda_{lap}\mathcal{L}_{lap} +
    \lambda_{int}\mathcal{L}_{int}
    \label{eq:lap_cons}
\end{equation}
where $\mathbf{n}^* = \{\mathrm{n}_i \mid i \in \mathcal{V} \}$ represents the estimated surface normals across all vertices. The weights $\lambda_{lap}$ and $\lambda_{int}$ balance the respective loss terms. Importantly, the Laplacian term does not act as an ad-hoc smoothness prior. It encodes a geometry-induced constraint derived from the discrete Laplacian, while the integrability term is introduced solely to ensure gradient consistency during optimization.
We solve this optimization problem using gradient descent to iteratively refine the estimates. The resulting $\mathbf{n}^*$ provide accurate surface normal estimates that satisfy both the shading and Laplacian constraints.

\subsection{Discussion on Limitations of Theory}

In this section, we discuss the limitations of our theory. 

\noindent \textbf{Case 1:} Our light source estimation using reverse photometric stereo requires at least three regions with sufficiently distinct local normal orientation. 
In the extreme case when the shape is simply a plane, we cannot estimate the light source direction and consequently cannot unambiguously estimate shape. 
However, if the light source direction is provided, even a simple plane can be unambiguously reconstructed. 

\noindent \textbf{Case 2:} In the special case of fronto-parallel lighting, the surface normals for both concave and convex shapes lie within the shading cone at all points on the surface and the ambiguities cannot be resolved.

\noindent \textbf{Case 3:} For directional lighting, surface normals along the line $p = -\frac{q_s}{p_s} q$ and their flipped counterparts produce the same shading and Laplacian weights. In the extreme case when the shape only contains normals along this line, we cannot resolve the ambiguity. 
Also note that, if all normals in a shape are collinear in the $pq$ space, 
the light source can also not be estimated as in Case 1. 
However, for general shapes, a nearby pixel would have a normal outside the above line. In this case, filling out based on nearby pixels allows selecting the correct sign of the normal.

Outside these special cases, the combination of shading and Laplacians provides a powerful way to resolve inherent shape ambiguities in each modality. Recall that, for unknown light source direction, a global binary ambiguity still exists. However, our approach estimates both possible light source direction and its corresponding shape. 

We note that the regular six-neighbor mesh was chosen for symmetry and computational convenience, and the derived equations in Sec.~\ref{subsec:L_to_pq} are not restricted to this configuration -- it can be extended to other mesh connectivities.

\section{Simulation Results}
We evaluate the theory and effectiveness of our proposed method through simulations on a variety of objects and under varying levels of input noise. In Sec.~\ref{subsec:source_est}, we analyze our light source estimation algorithm. Next, in Sec.~\ref{subsec:shape_est}, we compare our shape estimation algorithm against three baselines: (A) Shape from Laplacian, which optimizes $(p_i, q_i)$ to match the given Laplacian weights, (B) Shape from Shading, which optimizes $(p_i, q_i)$ to the match the given shading assuming known source direction, and (C) Analysis-by-synthesis, which optimizes for both $(p_s, q_s)$ and $(p_i, q_i)$ using the sum of Laplacian error and Shading error as loss. In Sec.~\ref{subsec:noise_analysis}, we analyze the robustness of our algorithm to noise. Finally, in Sec.~\ref{subsec:model_complexities}, we analyze the performance of our algorithm in the presence of inter-reflections, complex BRDFs, and spatially varying albedo. 

\begin{figure*}[t]
    \includegraphics[width=\textwidth, trim={1.1cm 1.9cm 2.0cm 1.5cm}, clip]{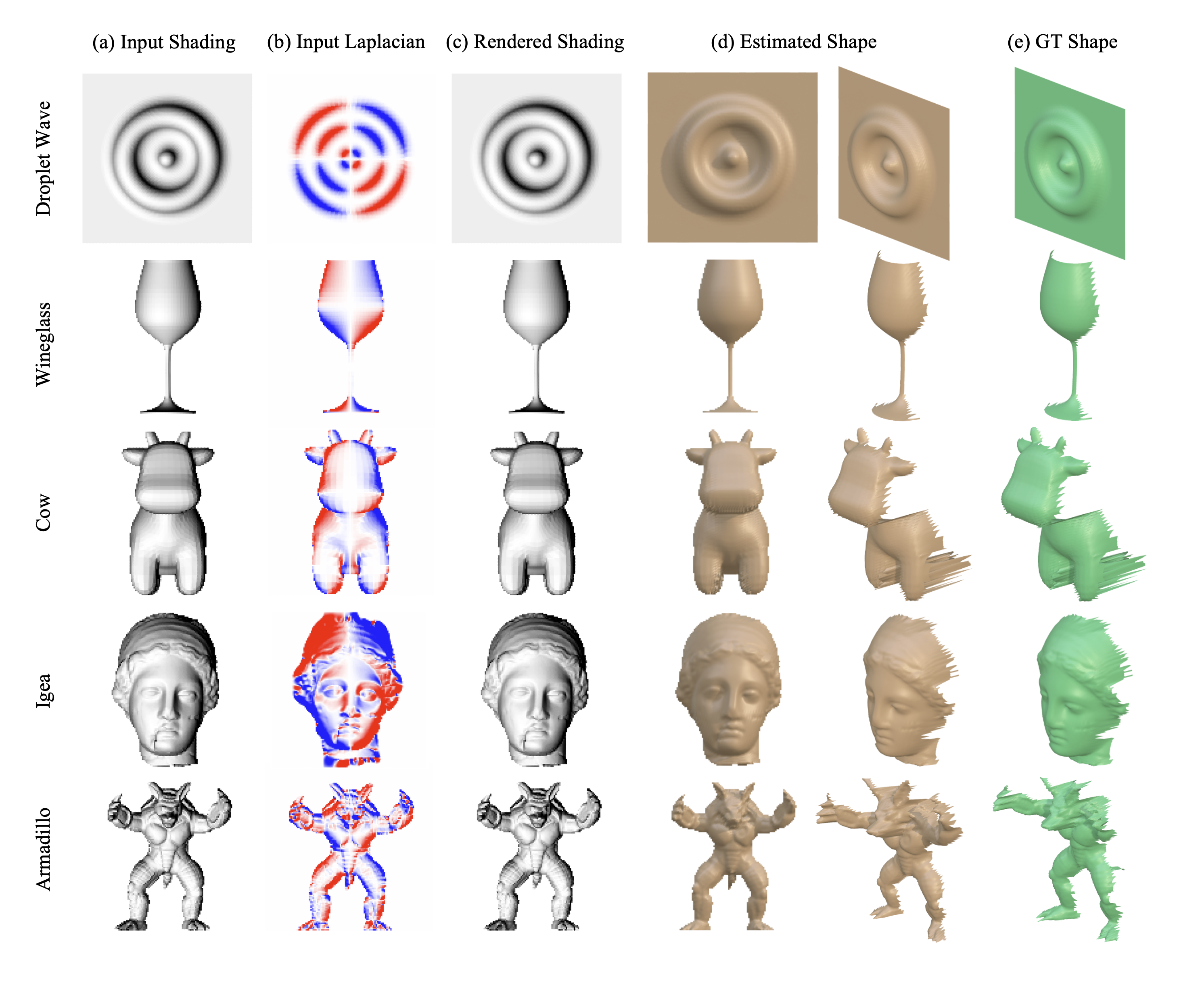}
        \caption{
        Shape estimation results for a simulated object using shading and Laplacian as inputs. Figures \textbf{(a)} and \textbf{(b)} show the inputs cues: shading and Laplacian, respectively. The Laplacian is visualized as the magnitude of Gaussian curvature, which is an intrinsic geometric property of shape. Figure \textbf{(c)} presents the rendering of the estimated lighting and reconstructed surface according to the algorithm described in Sec. \ref{sec:joint_est}. Note that this rendering is derived entirely from the unknown lighting and unknown shape. Figure \textbf{(d)} shows the depth map recovered from the reconstructed surface, and it can be seen that it aligns with the ground truth in \textbf{(e)}. The estimated shape and ground truth are both scaled, and the mean distance between them was at most 0.45\%. Detailed quantitative results are summarized in Table \ref{tab:sim_main}.}
    \label{fig:results_main}
    \vspace{-\intextsep}
\end{figure*}
\begin{figure}
    \includegraphics[width=\linewidth, trim={1.2cm 0.68cm 2.3cm 2.0cm}, clip]{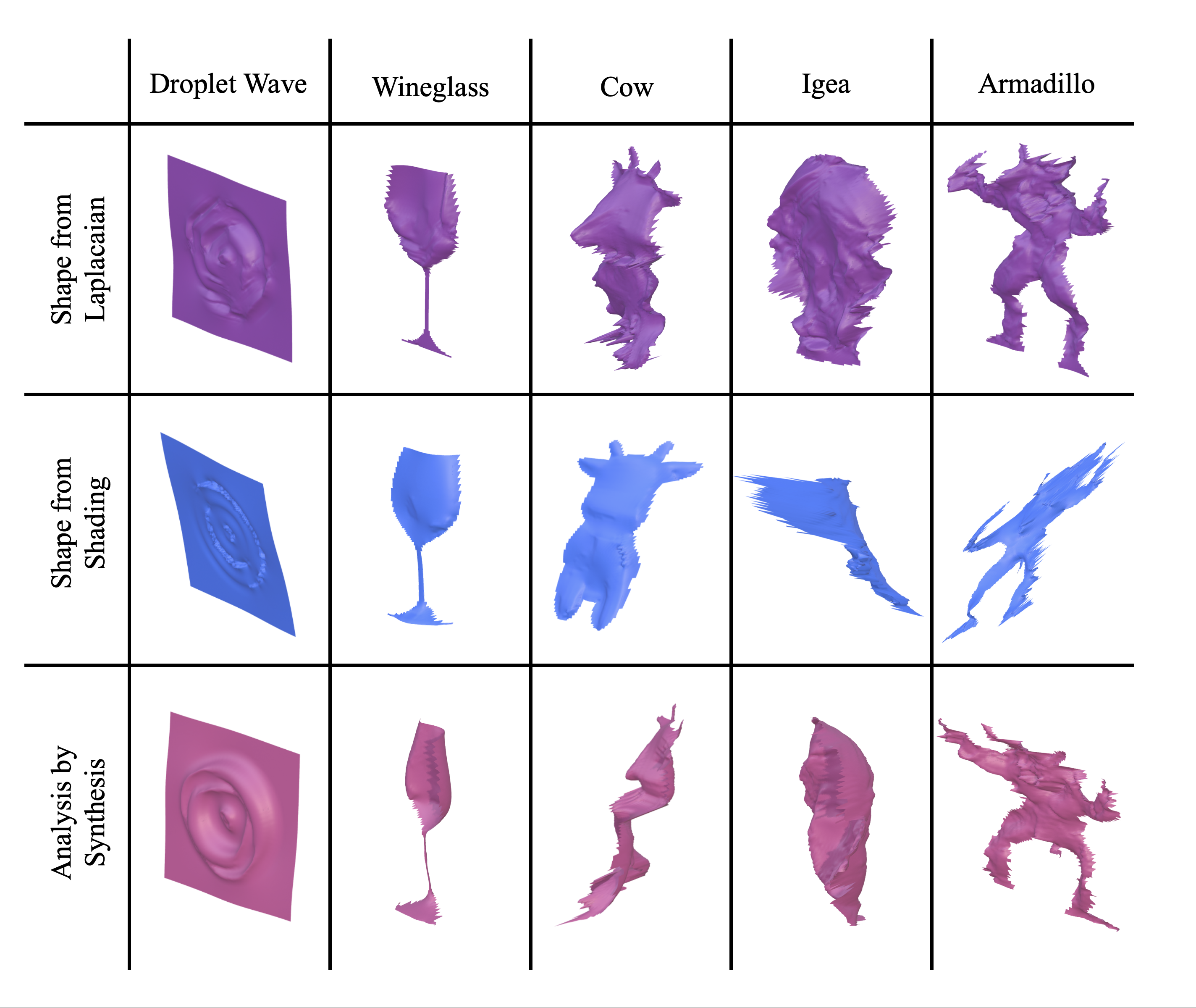}
        \caption{
        Shape estimation results using only Laplacian or shading cues, and joint optimization over lighting and shape using an analysis-by-synthesis approach. Due to the non-convex nature of the underlying optimization, the results exhibit ambiguities—locally binary ambiguities from Laplacian constraints and conic ambiguities from shading cues—reflecting the presence of multiple global minima. The light source estimation errors with the analysis-by-synthesis approach were $23.66^{\circ}$, $25.60^{\circ}$, $26.26^{\circ}$, $15.12^{\circ}$, and $21.34^{\circ}$ from left to right object, compared to our approach shown in Table \ref{tab:source_ang_error}.
        }
    \label{fig:results_baseline}
    \vspace{-\intextsep}
\end{figure}
\begin{figure}
    \includegraphics[width=\linewidth, trim={7.0cm 2.4cm 6.0cm 2.5cm}, clip]{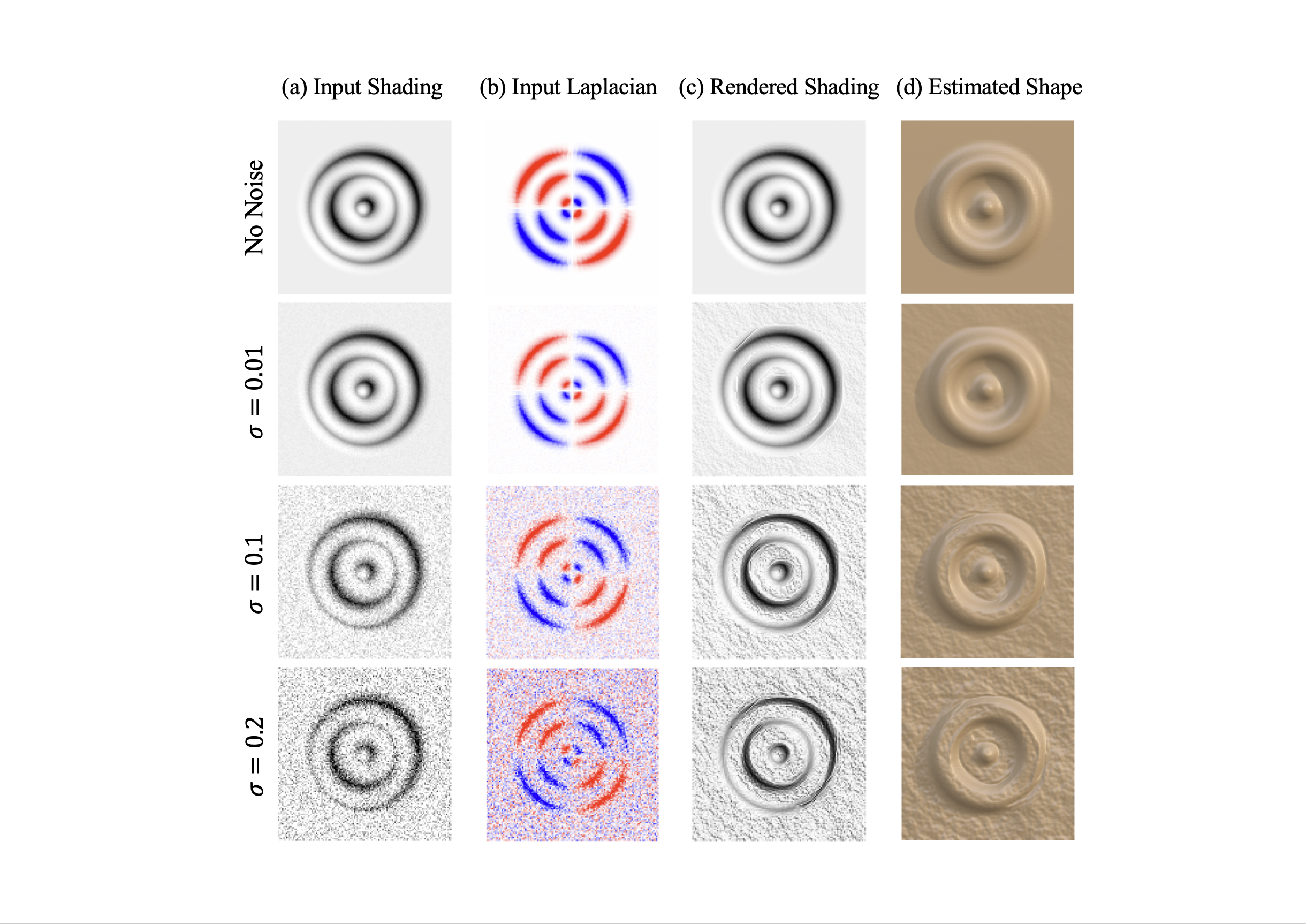}
        \caption{
        Shape estimation with varying levels of additive gaussian noise. \textbf{(a)} and \textbf{(b)} are the input shading and Laplacian, \textbf{(c)} the rendered shading image estimated from lighting and shape, and \textbf{(d)} the estimated depth-map. From the top to bottom, results are shown for no noise, and for Gaussian noise with standard deviations $\sigma=0.01$, $\sigma=0.1$, and $\sigma=0.2$, respectively. Although greater noise levels introduce more local distortion in the reconstructed shape, our proposed method still yields a reasonable estimation overall. Table \ref{tab:noise_results} presents the corresponding quantitative numbers.}
        \vspace{-\intextsep}
    \label{fig:noise_results}
\end{figure}

\begin{table}
    \centering
    \caption{
    Angular error (in degrees) between estimated and true lighting direction, evaluated after a single iteration of our refinement strategy and after convergence.}
        \resizebox{\linewidth}{!}{
            \begin{tabular}{cccccc}
                \toprule
                 & \textbf{Droplet Wave} & \textbf{Wineglass} & \textbf{Cow} & \textbf{Igea} & \textbf{Armadillo} \\
                \midrule
                \textbf{Initial Estimate}  & 27.02 & 2.026 & 1.076 & 48.94 & 25.03\\
                \textbf{After Refinement}  & 0.088 & 0.029 & 0.066 & 0.038 & 0.049\\
                \bottomrule
            \end{tabular}
        }
\label{tab:source_ang_error}
\end{table}

\begin{table}
    \centering
    \caption{
    Average percentage distance error with the ground truth for shapes estimated using shape from Laplacian\cite{Narayanan_2024_ECCV}, shape from shading\cite{frankot1988method}  (Fig. \ref{fig:results_baseline}), and proposed shape from Laplacian and shading (Fig. \ref{fig:results_main}), computed after scaling each shape to fit within a unit bounding box.}
        \resizebox{\linewidth}{!}{
            \begin{tabular}{cccccc}
                \toprule
                & \textbf{Droplet Wave} & \textbf{Wineglass} & \textbf{Cow} & \textbf{Igea} & \textbf{Armadillo} \\
                \midrule
                \textbf{Shape from Laplacian} & 2.64\% & 12.80\% &  3.23\% &  4.89\% &  5.56\% \\
                \textbf{Shape from Shading}   & 3.52\% & 12.97\% &  6.89\% & 12.21\% & 71.94\% \\
                \textbf{Analysis by Synthesis}& 2.23\% & 13.79\% & 13.65\% &  3.29\% & 28.08\% \\
                \textbf{SFLS (Ours)}          & 0.02\% &  0.06\% &  0.08\% &  0.06\% &  0.45\% \\
                \bottomrule
            \end{tabular}
        }
\label{tab:sim_main}
\end{table}
\begin{table}
    \centering
    \caption{
    Quantitative results for a droplet wave object with increasing Gaussian noise applied to the inputs of both the shading and Laplacian weights.}
    \begin{tabular}{lccc}
        \toprule
        Added Gaussian Noise         & $\sigma = 0.01$     & $\sigma = 0.1$      & $\sigma = 0.2$     \\
        \midrule
        \textbf{Avg. distance error} & $5\mathrm{e}^{-3}$  & $8\mathrm{e}^{-3}$  & $1\mathrm{e}^{-2}$ \\
        \textbf{\% Distance error}   & 0.49\%              & 0.79\%              & 1.12\%             \\
        \bottomrule
    \end{tabular}
\label{tab:noise_results}
\end{table}


\begin{table}
    \centering
    \caption{
    Average percentage distance error with the droplet wave object reconstructed from simulated noisy thermal videos with Gaussian noise and Gaussian + fixed pattern noise that is commonly seen in thermal cameras.}
    \begin{tabular}{lcc}
        \toprule
        & Gaussian & Gaussian + Fixed Pattern \\
        \midrule
        \textbf{SFLS with propagation \cite{Oharazawa_2025_ICCP}} & 0.81\%  & 1.20\%  \\
        \textbf{SFLS (Ours)}                                      & 0.36\%  & 0.42\%  \\
        \bottomrule
    \end{tabular}
\label{tab:noise_results_thermal}
\end{table}


\begin{figure}
    \centering
    \includegraphics[width=1.0\linewidth, trim={4.8cm 6.5cm 7.0cm 2.7cm}, clip]{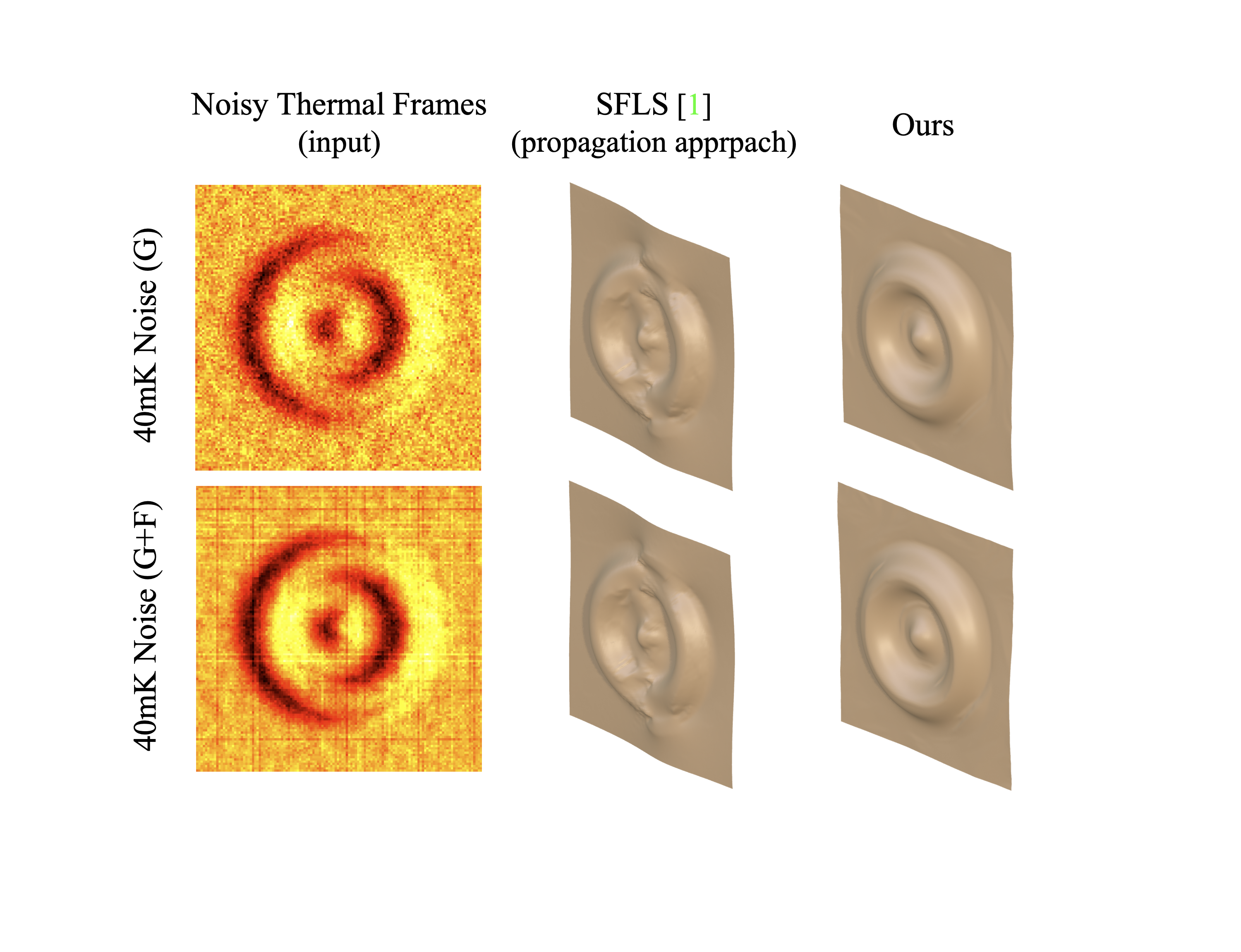}
        \caption{
        We simulated and reconstructed shape from the noisy thermal videos with Gaussian noise (G) and the Gaussian + fixed pattern noises (G+F). We used the algorithm from \cite{Narayanan_2024_ECCV} to estimate Shape Laplacian and we apply median filtering and total variation denoising before using them as input to our shape estimation. We compare our method with the propagation-based approach proposed in prior work \cite{Oharazawa_2025_ICCP}. Table \ref{tab:noise_results_thermal} presents the corresponding quantitative numbers.}
    \label{fig:noise_results_thermal}
\end{figure}
\begin{figure*}
    \includegraphics[width=1.0\linewidth, trim={0.8cm 0.0cm 0.5cm 0.0cm}, clip]{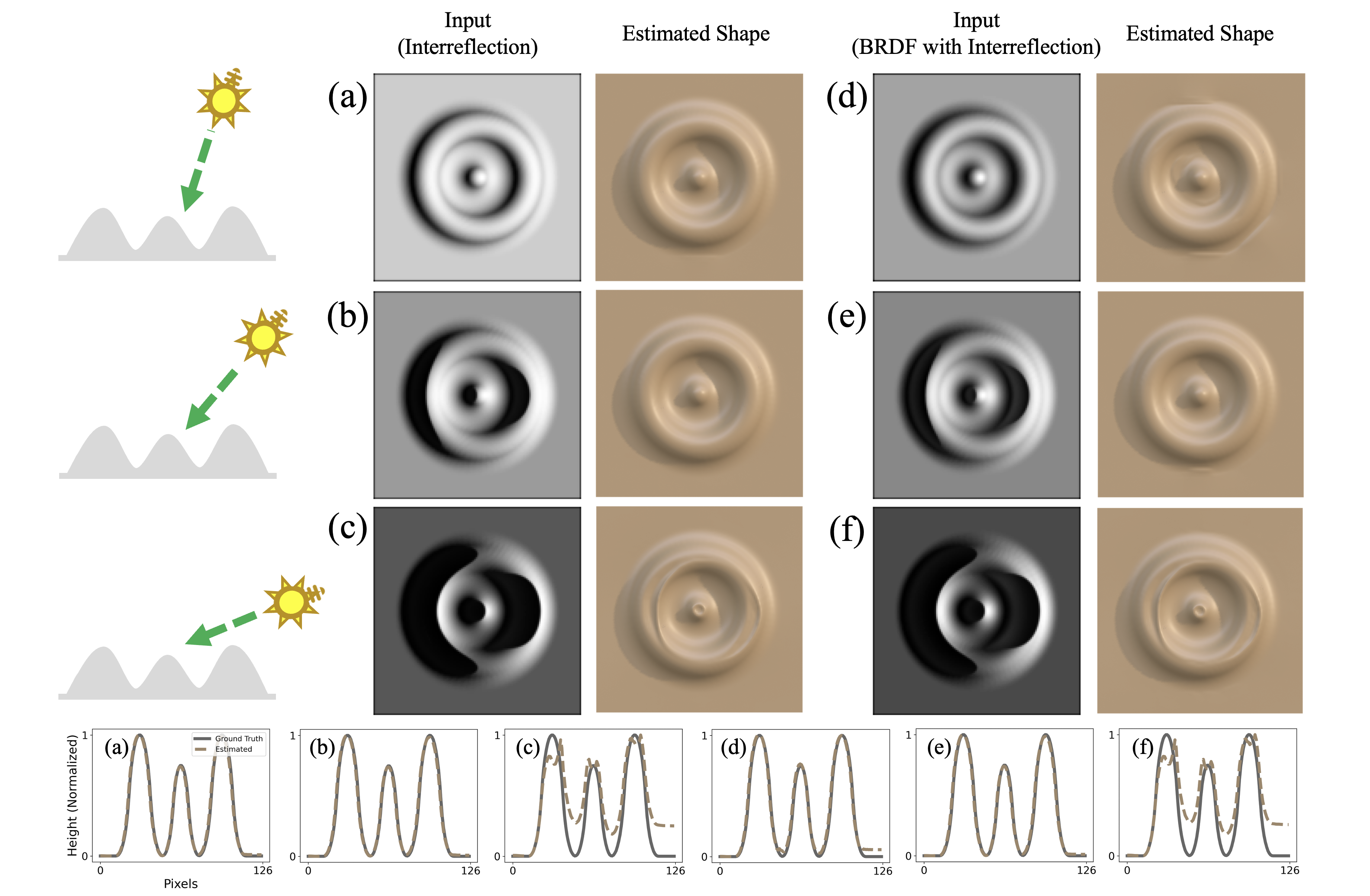}
    \caption{
    Shape estimation results for a simulated object under known lighting conditions, using Laplacian and shading as inputs for two cases: (1) shading includes interreflections with the Lambertian BRDF and (2) shading includes interreflections with a non-lambertian BRDF with roughness parameter $\alpha=0.4$. Figures \textbf{(a)}, \textbf{(b)} and \textbf{(c)} show pairs of the input shading image rendered using Mitsuba and the estimated shape. Similarly, Figures \textbf{(d)}, \textbf{(e)} and \textbf{(f)} show the corresponding results for the non-Lambertian BRDF case. Below the visual results, we also show the corresponding depth profiles comparing the estimated shapes with the ground truth along the vertical center line of the image. The mean distance between them was at most 0.48\%. Detailed quantitative results are in Table \ref{tab:analysis_interreflections}.}
    \label{fig:analysis_interreflections}
\end{figure*}
\begin{figure}[t]
    \includegraphics[width=\linewidth, trim={4.5cm 3.2cm 6.0cm 3.5cm}, clip]{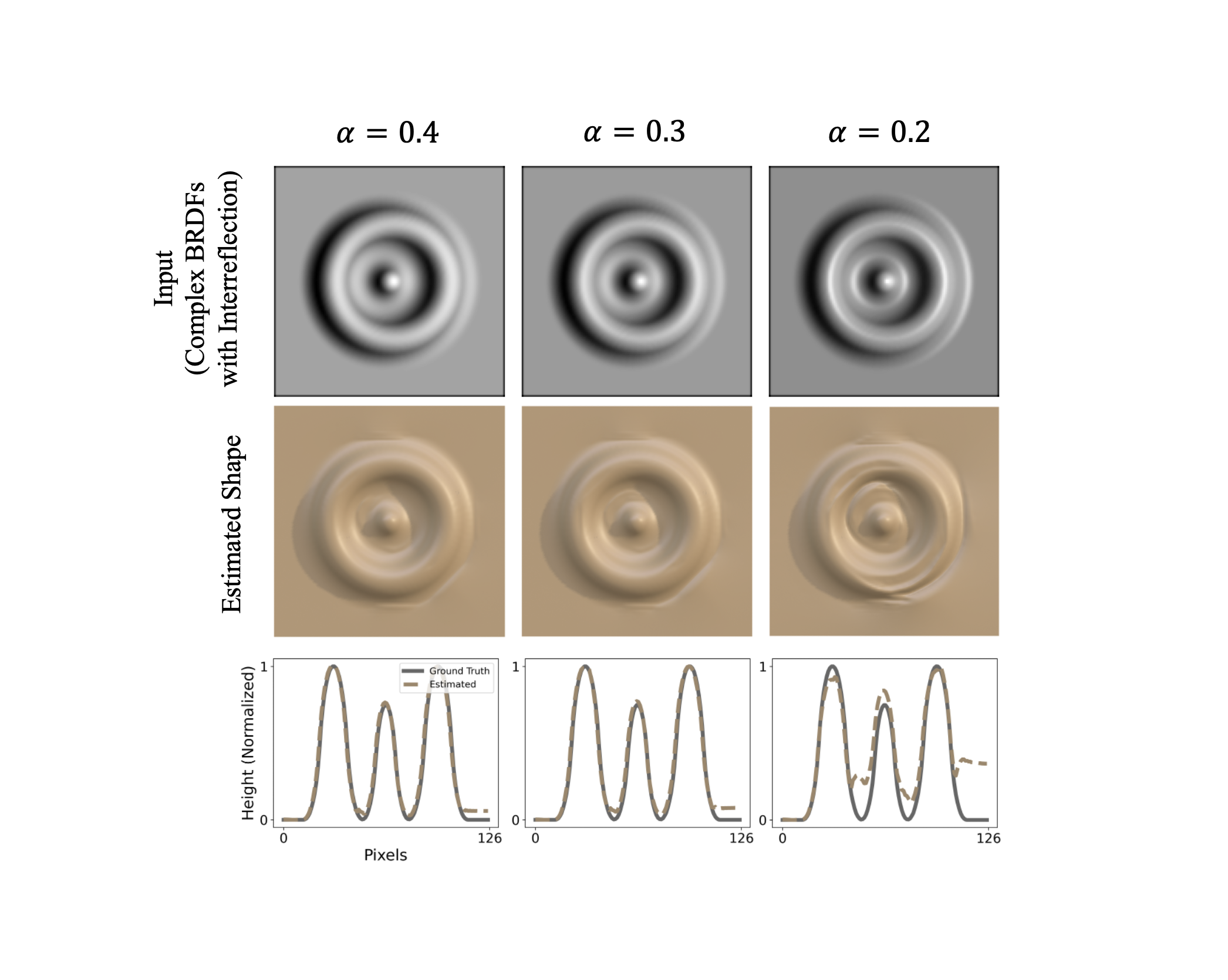}
    \caption{Shape estimation results using shading images with interreflections under different roughness parameters as input. As the surface becomes smoother (i.e., as $\alpha$ decreases), the specular highlights in the shading introduces rapid changes in surface normals. The depth profiles comparing the estimated shapes with the ground truth along the vertical center line of the image. Quantitative results are summarized in Table \ref{tab:analysis_brdfs}.}
    \label{fig:analysis_brdfs}
    \vspace{-\intextsep}
\end{figure}
\begin{figure}[t]
    \includegraphics[width=\linewidth, trim={12.8cm 11.19cm 16.7cm 8.0cm}, clip]{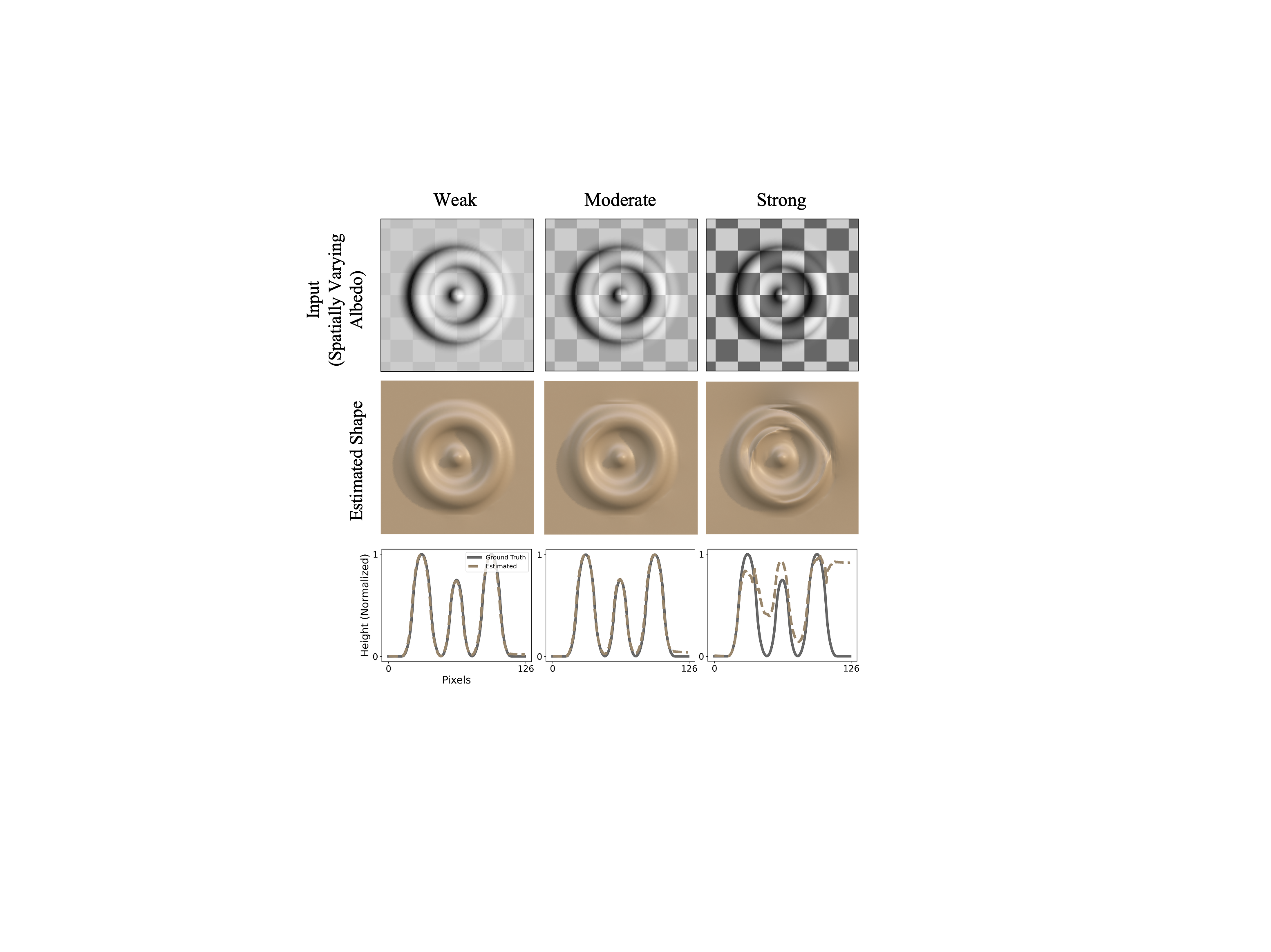}
    \caption{
    Shape estimation results using shading images with spatially varying albedo modeled using a checkerboard pattern. As the relative albedo contrast increases, high-contrast intensity variations increasingly confound shading cues and affect normal estimation. The depth profiles comparing the estimated shapes with the ground truth along the vertical center line of the image. Quantitative results are summarized in Table \ref{tab:analysis_albedo}.}
    \label{fig:analysis_albedo}
    \vspace{-\intextsep}
\end{figure}

\begin{table}
    \centering
    \caption{
    Quantitative results for a droplet wave for two cases (1) interreflections on a diffuse surface and (2) interreflections on a surface with complex BRDF, with roughness parameter $\alpha=0.4$.}
    \begin{tabular}{lccc}
        \toprule
        & $\theta=30$ & $\theta=50$ & $\theta=70$ \\
        \midrule
        \textbf{Interreflections}              & 0.02\% & 0.02\% & 0.43\% \\
        \textbf{BRDF with interreflections}    & 0.04\% & 0.02\% & 0.48\% \\
        \bottomrule
    \end{tabular}
\label{tab:analysis_interreflections}
\end{table}

\begin{table}
    \centering
    \caption{
    Quantitative results for a droplet wave object reconstructed from simulated BRDFs with interreflections. The lighting was set at $\theta=30$.}
        \begin{tabular}{lccc}
            \toprule
            & $\alpha = 0.4$ & $\alpha = 0.3$ & $\alpha = 0.2$ \\
            \midrule
            \textbf{BRDFs with interreflections}    & 0.09\% & 0.10\% & 0.57\% \\
            \bottomrule
        \end{tabular}
\label{tab:analysis_brdfs}
\end{table}

\begin{table}
    \centering
    \caption{
    Quantitative results for a droplet wave object reconstructed from simulated spatially varying albedo modeled using a checkerboard pattern. The lighting was set at $\theta=30$.}
        \begin{tabular}{lccc}
            \toprule
            & Weak & Moderate & Strong \\
            \midrule
            \textbf{Spatially varying albedo}    & 0.03\% & 0.06\% & 1.31\% \\
            \bottomrule
        \end{tabular}
\label{tab:analysis_albedo}
\end{table}

\begin{figure}
    \centering
    \includegraphics[width=0.75\linewidth, trim={13.5cm 12.0cm 12.5cm 5.5cm}, clip]{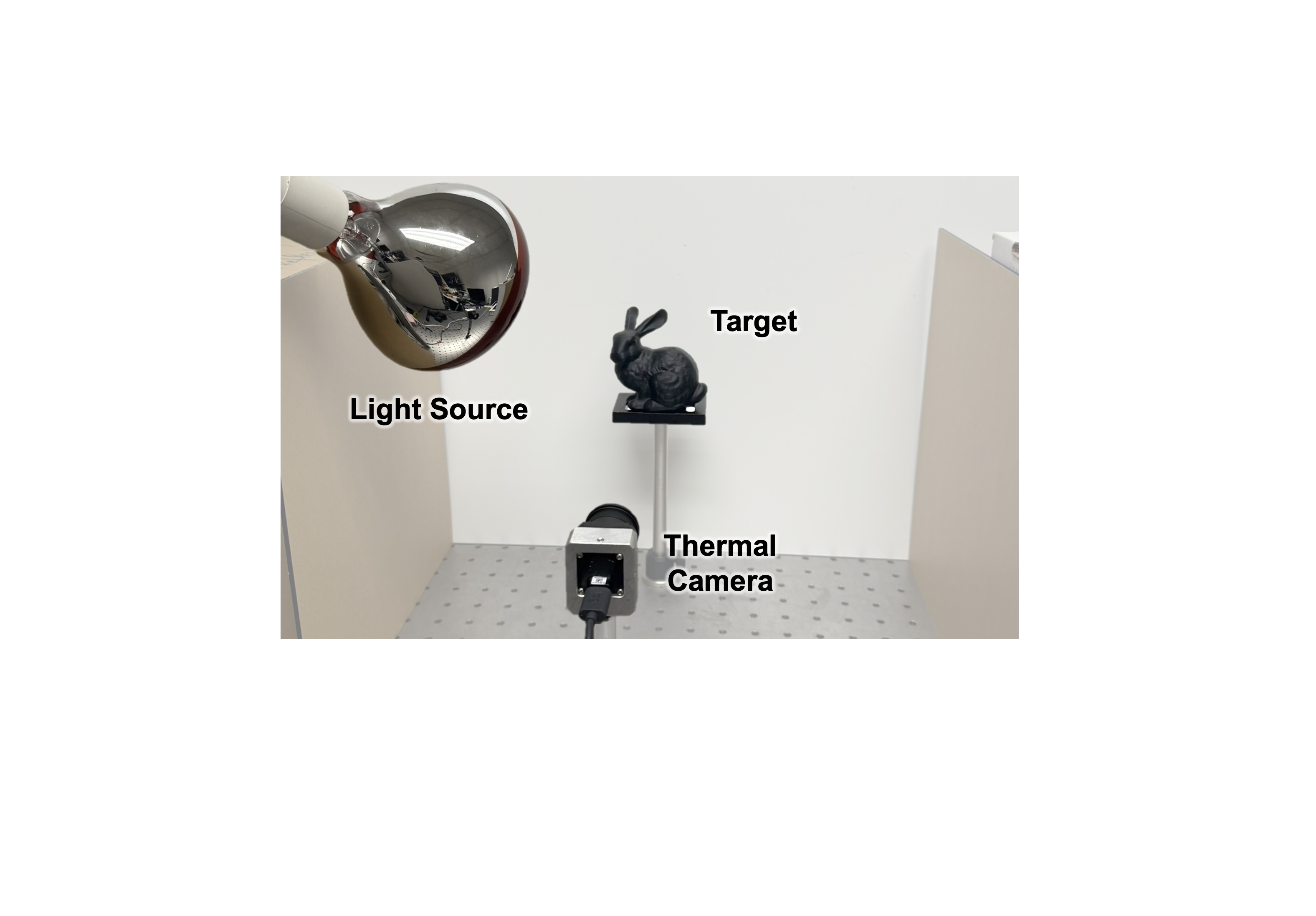}
        \caption{Our imaging system consists of a thermal camera and a light source. The light source is placed close to the object so that heat conduction induced by light absorption can be detected by the thermal camera.}
    \label{fig:results_real_ex_pic}
    \vspace{-\intextsep}
\end{figure}
\begin{figure*}
    \includegraphics[width=\linewidth, trim={2.2cm 4.5cm 2.5cm 1.8cm}, clip]{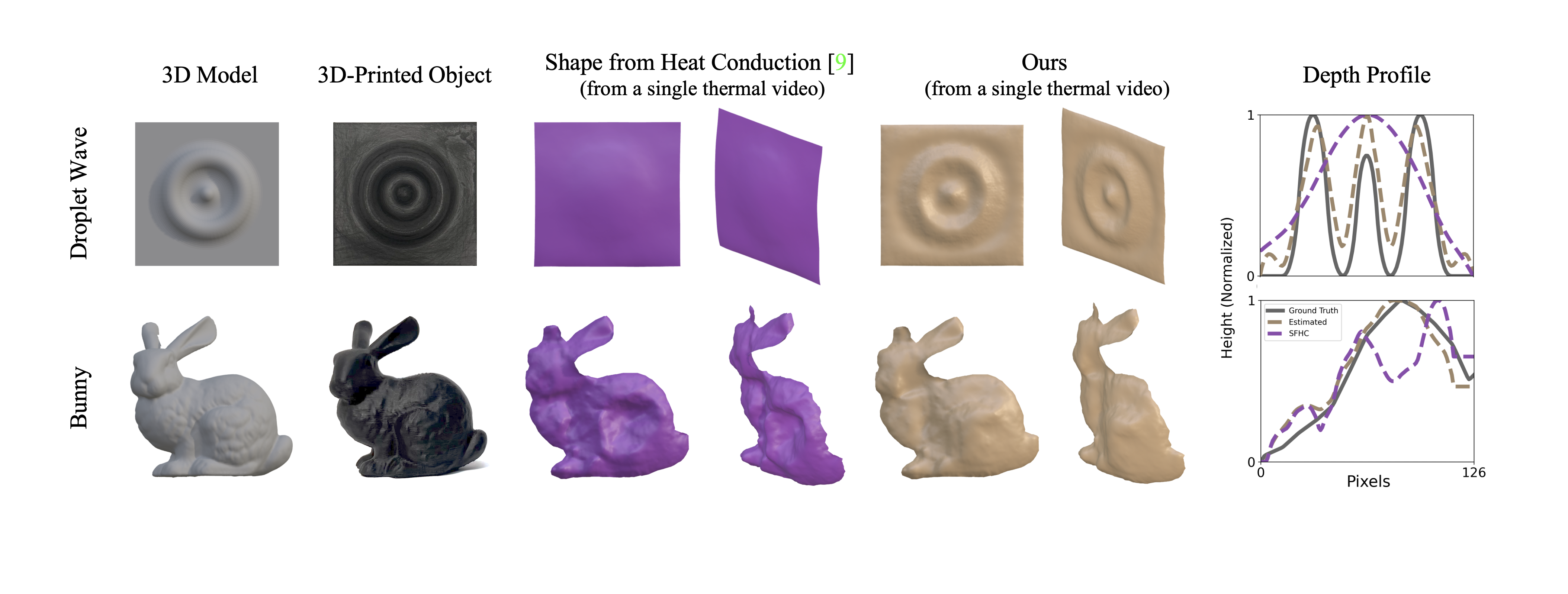}
        \caption{Shape estimation results from a single thermal video of a 3D-printed object using shape-from-heat-conduction \cite{Narayanan_2024_ECCV} and the proposed SFLS. From left to right, the reference 3D model, the captured image of the 3D-printed object, the shape estimated using shape-from-heat-conduction, and the shape estimated using SFLS. The depth profiles compare the estimated shapes with the ground truth along the vertical center line of the object. Detailed quantitative results are in Table \ref{tab:real_distance_3d_printed}.}
    \label{fig:results_real_3d_printed}
    \vspace{-\intextsep}
\end{figure*}
\begin{figure*}
    \includegraphics[width=\linewidth, trim={0.4cm 0cm 0.2cm 0.6cm}, clip]{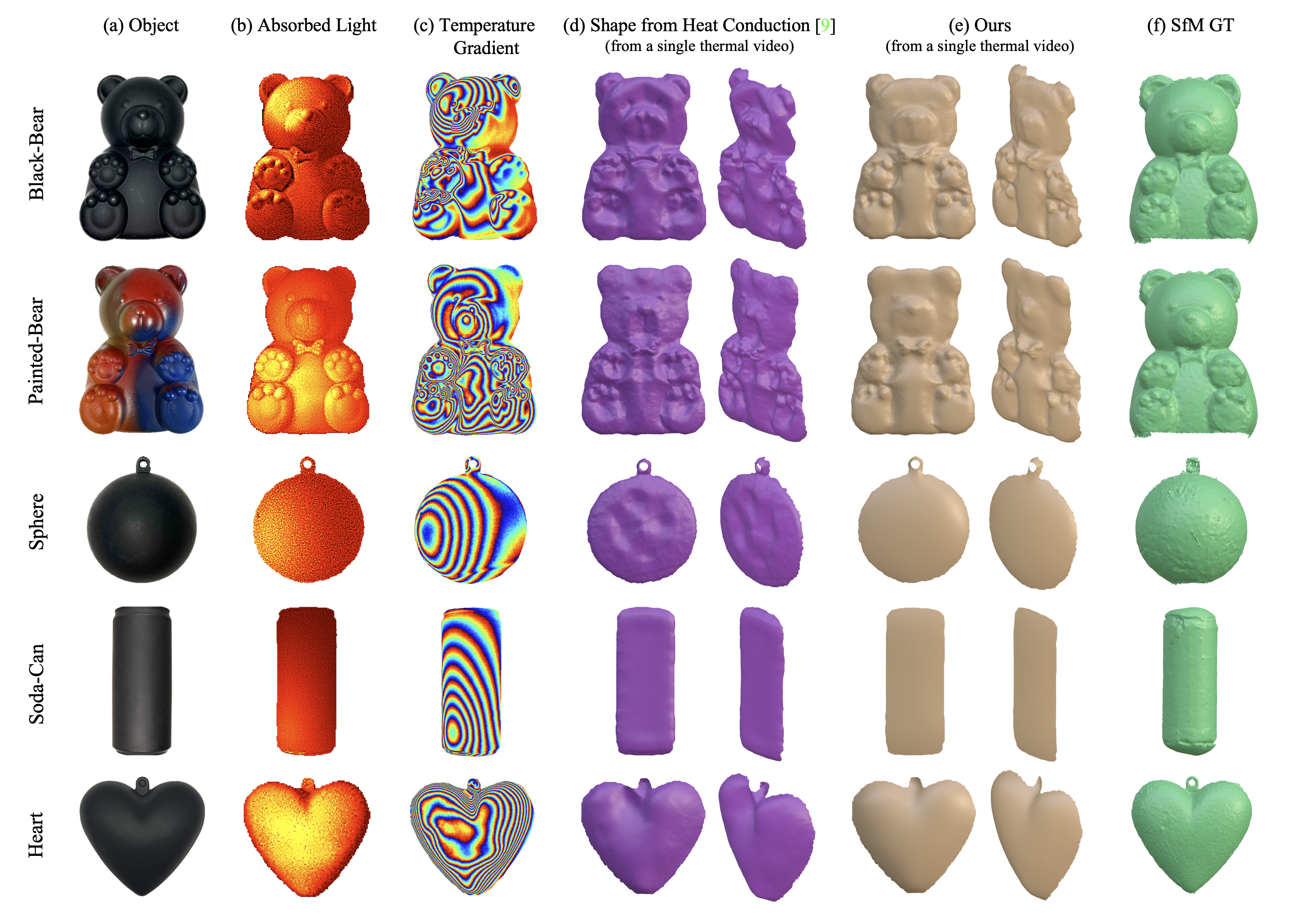}
        \caption{Shape estimation results from a single thermal video of real-world objects. Figure \textbf{(a)} show the real-world objects. Figure \textbf{(b)} show the absorbed light estimated from the thermal video, which is used as the shading input. Figure \textbf{(c)} show the corresponding temperature gradients. Figures \textbf{(d)} and \textbf{(e)} present the shapes estimated from shape-from-heat-conduction \cite{Narayanan_2024_ECCV} and the proposed SFLS. Figure \textbf{(f)} show the ground truth shapes obtained by SfM. Detailed quantitative results are summarized in Table \ref{tab:real_distance_sfm}.}
    \label{fig:results_real_comparison}
    \vspace{-\intextsep}
\end{figure*}

\begin{table}
    \centering
    \caption{
    Average percentage distance error with the ground truth 3D-model for shapes estimated using shape-from-heat-conduction \cite{Narayanan_2024_ECCV} with one thermal video, and the proposed shape from Laplacian and shading (Fig. \ref{fig:results_real_3d_printed}), computed after scaling each shape to fit within a unit bounding box.}
        \begin{tabular}{ccc}
            \toprule
            & \textbf{Droplet Wave} & \textbf{Bunny} \\
            \midrule
            \textbf{Shape from Heat Conduction \cite{Narayanan_2024_ECCV}} & 1.45\%  & 5.72\%  \\
            \textbf{SFLS (Ours)}                                           & 0.05\%  & 0.12\%  \\
            \bottomrule
        \end{tabular}
        \vspace{-\intextsep}
\label{tab:real_distance_3d_printed}
\end{table}

\begin{table}
    \centering
    \caption{
    Average percentage distance error with the ground truth SFM 3D-scans for shapes estimated using shape from heat conduction \cite{Narayanan_2024_ECCV} with one thermal video, and the proposed shape from Laplacian and shading (Fig. \ref{fig:results_real_comparison}), computed after scaling each shape to fit within a unit bounding box.}
        \resizebox{\linewidth}{!}{
            \begin{tabular}{cccccc}
            \toprule
             & \textbf{Black-Bear} & \textbf{Painted-Bear} & \textbf{Sphere} & \textbf{Soda-can} & \textbf{Heart} \\
            \midrule
            \textbf{Shape from Heat Conduction \cite{Narayanan_2024_ECCV}} & 3.06\% & 2.61\% & 3.43\% & 2.44\% & 2.13\% \\
            \textbf{SFLS (Ours)}                                           & 0.23\% & 1.53\% & 0.97\% & 0.74\% & 1.05\% \\
            \bottomrule
            \end{tabular}
        }
\label{tab:real_distance_sfm}
\end{table}


\subsection{Source Estimation Results}
\label{subsec:source_est}

Table \ref{tab:source_ang_error} presents the angular error between the estimated and ground truth light source directions across various simulated objects. 
Initial estimates based on near-planarity-based analytical normal initialization result in errors as high as $49^{\circ}$ for some cases. 
However, by iteratively refining the estimates through alternating reverse photometric stereo and Laplacian-constrained normal refinement, the angular error is reduced dramatically to below $0.09^{\circ}$. 

\subsection{Shape Estimation Results}
\label{subsec:shape_est}

Figure \ref{fig:results_baseline} shows the qualitative results of shape estimation using the three baselines. Shape from Laplacian estimates the correct local curvature while the overall shape is incorrect due to local concave/convex ambiguities. On the other hand, Shape from Shading estimates shapes that often have lower curvature than the true shape. 
The analysis-by-synthesis approach, i.e. joint optimization over lighting and shape, is simple in principle but gets stuck in local minima. Note that for every source direction, there exist many corresponding shapes that exactly satisfy the shading equation. Thus, fixing the source direction is crucial for converging to the correct solution. Unlike the shading equation, the Shape Laplacians at near-planar pixels can be solved in closed form, which we then use to estimate source. 
Also, joint optimization requires a hyperparameter to balance photometric error and Laplacian error, while our proposed sign selection and refinement satisfy the shading equation by construction.
The corresponding results from our proposed method is shown in Fig.~\ref{fig:results_main} along with the ground truth shape. By combining the correct curvature information from Shape Laplacians and the correct direction information from Shading, our algorithm faithfully recovers the 3D shapes even for complex geometries.

To quantitatively evaluate the performance, we compare the estimated meshes against ground truth meshes using MeshLab after global registration. Table \ref{tab:sim_main} reports the average percentage distance errors.
Our proposed Shape-from-Laplacian-and-Shading (SFLS) method achieves high reconstruction accuracy, with errors as low as 0.45\% for complex geometries such as the Armadillo object.

In addition, our implementation does not require the propagation-based shape estimation used in prior work \cite{Oharazawa_2025_ICCP}, and instead refines the entire shape simultaneously. As a result, while previous approaches require 2--3 hours per scene, the proposed method achieves shape estimation in only 1--2 minutes.

\subsection{Robustness to Noise}
\label{subsec:noise_analysis}

The inputs to our algorithm are the Shading image and Shape Laplacian weights. 
To evaluate robustness, we added Gaussian noise to both these inputs. Table \ref{tab:noise_results} shows the percentage distance errors for the reconstructed droplet shape under increasing noise levels. Despite significant input degradation, our method maintains errors below 1.2\%, demonstrating good resilience. Visual results in Fig. \ref{fig:noise_results} confirm the robustness of the proposed pipeline.

While additive gaussian noise is a reasonable noise model for Shading images, the noise model for Shape Laplacians could be more complex as they are estimated from thermal videos which suffer from both gaussian and fixed pattern noises. Therefore, we performed additional experiments by simulating thermal videos and estimating Shape Laplacians using the method in ~\cite{Narayanan_2024_ECCV}. We additionally use median filtering and total variation denoising on the estimated Shape Laplacians as pre-processing steps.

Fig. \ref{fig:noise_results_thermal} shows the reconstructed shapes from thermal video with added Gaussian noise, and Gaussian and fixed pattern noise. Table \ref{tab:noise_results_thermal} shows the distance errors for each case. The errors are within 0.5\% even under noise conditions commonly observed in real-world thermal imaging.

\subsection{Complex reflectance and illumination effects}
\label{subsec:model_complexities}

In this section, we analyze our algorithm in the presence of complex light transport effects such as shadows, inter-reflections, non-Lambertian BRDFs, and spatially varying albedo.
Our light source estimation algorithm based on reverse photometric stereo fails in most of these cases with errors above $40^\circ$. Therefore, we restrict our analysis to the case of known lighting. 

The behavior of our algorithm varies depending on the amount of interreflection present in the scene. In regions where the direct component (for Lambertian surfaces, this is the cosine shading) dominates interreflections, our approach performs reasonably even if it produces slightly shallower shapes (as noted in classical works \cite{DBLP:journals/ijcv/NayarIK91}). But when interreflections dominate, using just the cosine shading model can result in larger errors. In particular, in shadow regions where there are only interreflections, the surface patches may be estimated to be facing away from the source when they are actually oriented towards the source. Thus, in such cases, the binary ambiguities in Shape Laplacian will not be resolved correctly using our simple shading model.

To experimentally confirm these observations, we simulate the Droplet Wave shape under three different lighting conditions with interreflections rendered using Mitsuba 3~\cite{Mitsuba3}. We further analyze the same scenes with a non-Lambertian BRDF by varying the roughness ($\alpha$) parameter and the albedo spatially. 
Fig.~\ref{fig:analysis_interreflections} shows the corresponding shape estimation results.
When the lighting is close to fronto-parallel, most regions have a dominant direct component and our shape estimation resolves the concave/convex ambiguity.
As lighting goes toward lower polar angles, more regions are shadowed and the shape estimation progressively degrades for both Lambertian and non-Lambertian BRDFs. 
Table \ref{tab:analysis_interreflections} reports the corresponding quantitative errors for these experiments.  
Fig. \ref{fig:analysis_brdfs} compares the shape estimation results with varying roughness parameters. As shown in Table \ref{tab:analysis_brdfs}, the errors increase with lower roughness (more specular). 
The profile plots in the bottom row show that the specular regions in the shading introduces rapid changes in surface normals near specular highlights.
Fig. \ref{fig:analysis_albedo} shows the shape estimation results with spatially varying albedo modeled using a checkerboard pattern. As shown in Table \ref{tab:analysis_albedo}, the errors increase as the albedo contrast becomes higher. Specifically, for high-contrast albedo, intensity variations due to albedo dominate shading cues, resulting in the failure of our simple shading model to correctly resolve the binary ambiguities inherent in the Shape Laplacian.

\section{Real results}
\label{subsec:real_res}
In this section, we evaluated the effectiveness of our method using a variety of real-world objects with diverse shapes, including complex ridges and grooves, as shown in Figures \ref{fig:results_real_3d_printed} and \ref{fig:results_real_comparison}. For homogeneous albedo, shading is directly proportional to the absorbed light \cite{Ramanagopal_2024_CVPR}. This enables consistent integration of shading with Laplacian cues only using heat conduction without requiring a separate shading image. Based on this relationship, we model shading in our real-world experiments as a source term in the heat transport equation. We removed normal constraints from a Shape-from-Heat-Conduction-based optimization framework and explicitly integrated shading and Laplacian constraints to resolve the concave/convex ambiguity.

\subsection{Hardware setup}
We used a FLIR Boson thermal camera with a resolution of 512 $\times$ 640 (24$^\circ$ HFOV). A single light source was placed at an uncalibrated position, and a 15-second video was captured while the light source was turned on (see Fig. \ref{fig:results_real_ex_pic}). Prior to acquisition, a flat field correction (FFC) was performed and then disabled for the remainder of the capture. The first ten frames served to remove reflections and fixed pattern noise, and the remaining frames were processed by the proposed method at 10 FPS.

\subsection{Details of real-world objects}
The objects used in our experiments are made of different materials. The Droplet Wave object and the Bunny were fabricated by 3D printing using plastic filament based on 3D models (Fig. \ref{fig:results_real_3d_printed}). In addition to our own captured data, we also evaluated our method on the Shape-from-Heat-Conduction dataset~\cite{Narayanan_2024_ECCV}, which contains the Black bear, Sphere, and Soda can scenes. The Bear, Sphere, and Heart are made of acrylic plastic, while the Soda can is made of an aluminum alloy.
The objects shown in Fig. \ref{fig:results_real_comparison} were painted black to enhance light absorption, with approximately known emissivity. In addition, for the Painted bear, we applied several paints to evaluate the effectiveness of our method under varying visible reflectance conditions.

\subsection{Evaluation and Comparisons}
Figure \ref{fig:results_real_3d_printed} shows the estimated shapes of objects fabricated using 3D printing. The proposed method robustly reconstructs the overall object geometry, including complex surface details such as concave and convex structures. In particular, while existing Shape-from-Heat-Conduction–based approaches \cite{Narayanan_2024_ECCV} suffer from an inherent concave/convex ambiguity in the Laplacian estimated from the heat transport equation using a single thermal transient video, the proposed method effectively resolves this ambiguity and recovers the correct surface shape.
Table~\ref{tab:real_distance_3d_printed} reports the quantitative errors compared with respect to the ground truth 3D models.

Figure \ref{fig:results_real_comparison} shows reconstruction results on real-world objects under varying surface and reflectance conditions. Specifically, we compare reconstruction results for identically shaped bear objects, including one painted black and others painted with different reflectance properties and spatially varying albedo. Across all cases, the proposed method robustly recovers the global shape structure.
Table \ref{tab:real_distance_sfm} reports the distance errors for each object with respect to dense SfM scans. The errors remain within 1.6\% even under challenging conditions with spatially varying albedo. 
These results are consistent with those observed in the simulation experiments and demonstrate that our method remains effective under a wide range of visible BRDF conditions.

\section{Conclusion}
\label{sec:conclusion}

Heat conduction has recently emerged as a novel way to determine the Shape Laplace-Beltrami operator~\cite{Narayanan_2024_ECCV}, which is an intrinsic representation of shape, under unknown lighting and even for non-Lambertian surfaces.
However, recovering the explicit shape suffers from a local binary convex/concave ambiguity. 
In contrast, estimating the shape from a single shading image is a well-known ill-posed problem even for a Lambertian scene with uniform albedo and known source direction.
At each pixel, there is a cone of possible normals around the source direction that produce the same shading image. 
In this paper, we have shown that, outside special cases, only one of the convex/concave solutions to the Shape Laplacians lies on the Shading cone.

Our key insight is that the Laplacian weights can be directly computed from the vertex normals that are used in the shading equation, thus enabling a seamless integration of two complementary cues. 
Unlike Shape-from-Shading algorithms that use an arbitrary smoothness term, the shape Laplacian encodes the true curvature of the shape. 
As a result, combining shading and Laplacian constraints naturally produces a unique solution.
Extending our theory to more complex light transport (near-source lighting, inter-reflections, complex BRDFs) and perspective cameras is an interesting avenue for future work.

\section*{Acknowledgments}
The algorithm and the simulation results presented in this paper are based on and extended from our prior work \cite{Oharazawa_2025_ICCP} at 2025 IEEE International Conference on Computational Photography (ICCP). The research was supported by NSF Award IIS-2107236 and NSF-NIFA AI Institute for Resilient Agriculture (AIIRA). Kao Corporation supported Akihiko Oharazawa's effort. We sincerely thank Zeqing Leo Yuan for the insightful discussions.

 
\bibliographystyle{IEEEtran}
\bibliography{references}


\begin{IEEEbiography}[{\includegraphics[width=1in,height=1.25in,clip,keepaspectratio]{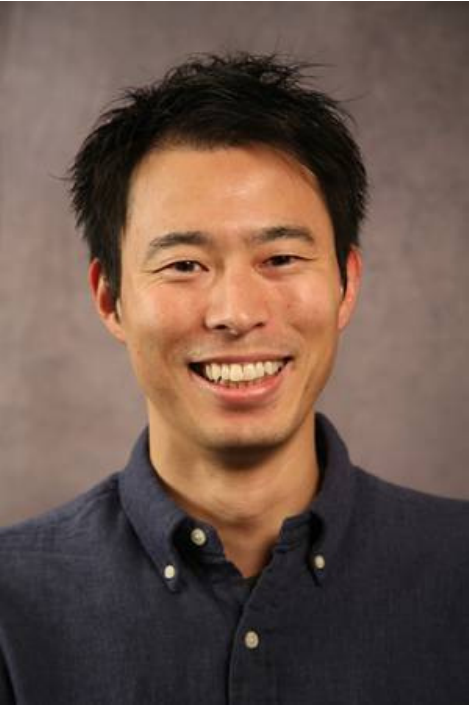}}]{Akihiko Oharazawa}
is a visiting researcher at the Robotics Institute, Carnegie Mellon University, where he works with Prof. Srinivas Narasimhan. He is also a Research Scientist at the Skincare Laboratory of Kao Corporation. He received his M.S. in Comprehensive Human Science from the University of Tsukuba. His current research interests lie in physics-based vision, computer vision, and computational imaging.
\end{IEEEbiography}

\vspace{11pt}
\begin{IEEEbiography}[{\includegraphics[width=1in,height=1.25in,clip,keepaspectratio]{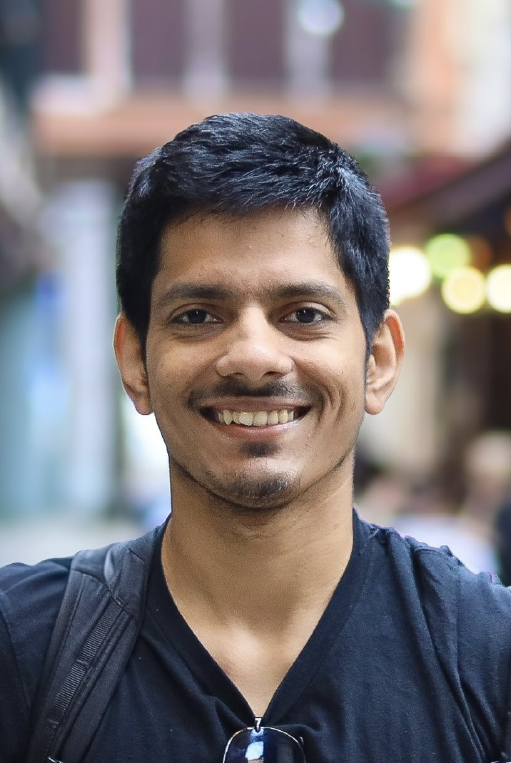}}]{Sriram Narayanan}
is a Ph.D. student in the Robotics Institute at Carnegie Mellon University, advised by Prof. Srinivasa Narasimhan. His research lies at the intersection of computer vision, computational imaging, and robotics, with a focus on physics-based methods for heat and light transport. Prior to CMU, he was a Research Scholar at NEC Laboratories America. He is a recipient of the Uber Presidential Fellowship.
\end{IEEEbiography}

\vspace{11pt}
\begin{IEEEbiography}[{\includegraphics[width=1in,height=1.25in,clip,keepaspectratio]{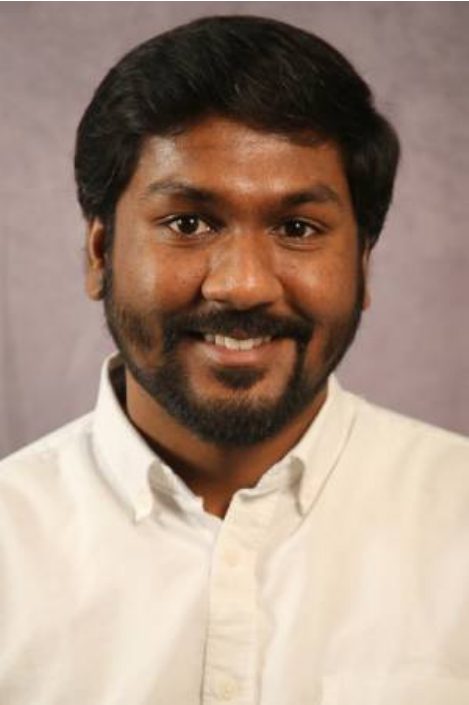}}]{Mani Ramanagopal}
is a Project Scientist at the Robotics Institute at Carnegie Mellon University. He obtained his Ph.D. in Robotics from the University of Michigan in 2022 and a B.Tech \& M.Tech in Electrical Engineering from Indian Institute of Technology Madras in 2016. He is the recipient of the Richard F. and Eleanor A. Towner Prize for Distinguished Academic Achievement for Outstanding Graduate Student. His research interests are in robotics, vision, and imaging, with a focus on physics-based approaches.
\end{IEEEbiography}

\vspace{11pt}
\begin{IEEEbiography}[{\includegraphics[width=1in,height=1.25in,clip,keepaspectratio]{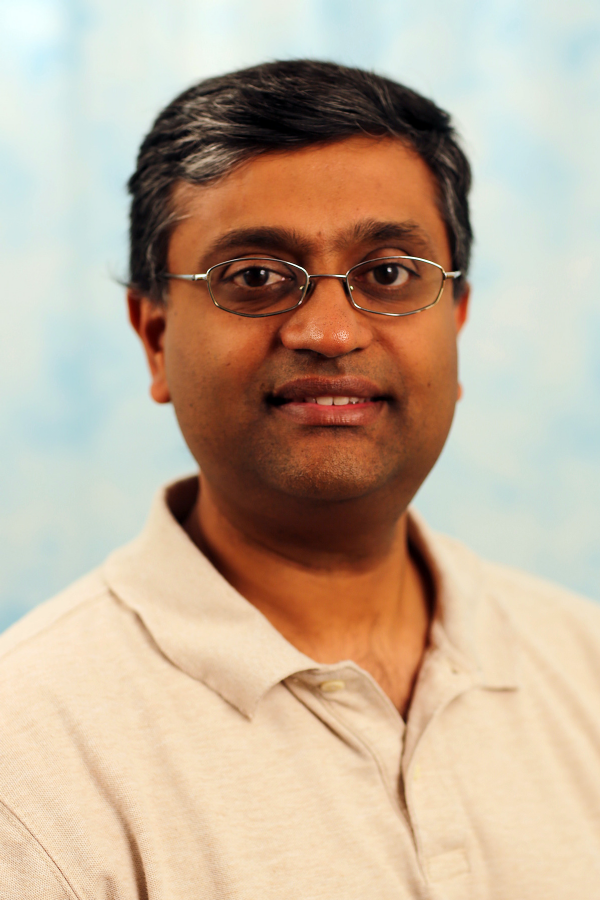}}]{Srinivasa G. Narasimhan}
is the Interim Director and U. A and Helen Whitaker Professor of the Robotics Institute at Carnegie Mellon University. He obtained his Ph.D. from Columbia University in Dec 2003. His group focuses on novel techniques for imaging and illumination to enable applications in vision, graphics, robotics, agriculture, intelligent transportation, and medical imaging. His works have received more than a dozen Best Paper or Best Demo or Honorable mention awards at major conferences. In addition, he has received the Ford URP Award (2013), Okawa Research Grant (2009) and the NSF CAREER Award (2007). He is the co-inventor of programmable headlights, Aqualux 3D display, Assorted pixels, Motion-aware cameras, Episcan360, Episcan3D, EpiToF3D, and programmable triangulation light curtains. He served on the editorial board of the International Journal of Computer Vision and serves as Senior or Lead Area Chair of top computer vision conferences. 
\end{IEEEbiography}



\vfill

\end{document}









\appendix
\noindent\textbf{Laplacian weights from vertex normals}
\label{sec:s1-laplacian}

\noindent
Here, we provide the detailed equations for the Laplacian weights for all six edges of a vertex. 
\begin{align}
w_{ia}  &= \frac{1}{2} \Bigg( \frac{p_b^2 + p_bq_a + 1}{ \sqrt{1 + q_a^2 + p_b^2}} + \frac{p_i^2 + p_iq_a + 1}{ \sqrt{1 + q_a^2 + p_i^2}} \Bigg) \notag \\
w_{ij}  &= \frac{1}{2} \Bigg( \frac{q_ap_i + q_a^2 + 1}{ \sqrt{1 + q_a^2 + p_i^2}} + \frac{q_jp_i + q_j^2 + 1}{ \sqrt{1 + q_j^2 + p_i^2}} \Bigg) \notag \\
w_{ik}  &= \frac{1}{2} \Bigg( \frac{-q_jp_i           }{ \sqrt{1 + q_j^2 + p_i^2}} - \frac{p_lq_i            }{ \sqrt{1 + p_l^2 + q_i^2}} \Bigg) \notag \\
w_{il}  &= \frac{1}{2} \Bigg( \frac{p_lq_i + p_l^2 + 1}{ \sqrt{1 + p_l^2 + q_i^2}} + \frac{p_cq_i + p_c^2 + 1}{ \sqrt{1 + p_c^2 + q_i^2}} \Bigg) \notag \\
w_{ic}  &= \frac{1}{2} \Bigg( \frac{q_i^2 + q_ip_c + 1}{ \sqrt{1 + p_c^2 + q_i^2}} + \frac{q_bp_c + q_b^2 + 1}{ \sqrt{1 + p_c^2 + q_b^2}} \Bigg) \notag \\
w_{ib}  &= \frac{1}{2} \Bigg( \frac{-p_cq_b           }{ \sqrt{1 + p_c^2 + q_b^2}} - \frac{p_bq_a            }{ \sqrt{1 + q_a^2 + p_b^2}} \Bigg)
\label{eq:cotan-weight-all}
\end{align}

\vspace{0.8em}
\noindent\textbf{Near-planar patch solutions}
\label{sec:s2-near-planar}

\noindent
Let $D = w_1 + w_2 + 2w_3$. The two solutions are
\begin{equation}
    x = \frac{D \pm \sqrt{D^2 - 4}}{2}.
\end{equation}
As the solutions need to be real, $D^2-4 \geq 0$. Since all terms in the definition of $D = w_1 + w_2 + 2w_3$ are positive, we have $D>0$. The above two inequalities imply $D\geq2$. 
We can then write
\begin{align}
    (D-2)^2 &< (D-2)(D+2) \nonumber \\ 
    D - 2 &< \sqrt{D^2 - 4} \nonumber \\
    D - \sqrt{D^2 - 4} &< 2  \nonumber \\
    \frac{D - \sqrt{D^2 - 4}}{2} &< 1
\end{align}
Since $D>2$, it is trivial to verify that
\begin{equation}
    x = \frac{D}{2} + \frac{\sqrt{D^2-4}}{2} \geq 1.
\end{equation}
When $D=2$, both solutions are identical ($x=1$).


 
\bibliographystyle{IEEEtran}
